\documentclass[final, 3p, times, twocolumn, authoryear]{elsarticle}

\usepackage{epsfig}
\usepackage{tabularx}
\usepackage{latexsym}
\usepackage{subcaption}
\usepackage{lineno}
\usepackage{amssymb}
\usepackage{amsmath}

\journal{Computers and Electronics in Agriculture}

\begin{document}

\begin{frontmatter}



\title{Multimodal surface defect detection from wooden logs for sawing optimization} 

\author[lut,but]{Bořek Reich} 
\ead{borek.reich@student.lut.fi}
\author[lut,but]{Matej Kunda} 
\ead{36matej36@gmail.com}
\author[lut]{Fedor Zolotarev} 
\ead{tedzolotarev@gmail.com}
\author[lut]{Tuomas Eerola} 
\ead{tuomas.eerola@lut.fi}
\author[lut,but]{Pavel Zemčík} 
\ead{pavel.zemcik@lut.fi}
\author[finnos]{Tomi Kauppi} 
\ead{tomi.kauppi@finnos.fi}

\affiliation[lut]{organization={Lappeenranta-Lahti University of Technology},
            addressline={Yliopistonkatu 34}, 
            city={Lappeenranta},
            postcode={53850}, 
            country={Finland}}

\affiliation[but]{organization={Brno University of Technology, Faculty of Information Technology},
            addressline={Božetěchova 1/2}, 
            city={Brno},
            postcode={61266}, 
            country={Czech Republic}}

\affiliation[finnos]{organization={Finnos Oy},
            addressline={Tukkikatu 5}, 
            city={Lappeenranta},
            postcode={53900}, 
            country={Finland}}

\begin{abstract}
We propose a novel, good-quality, and less demanding method for detecting knots on the surface of wooden logs using multimodal data fusion. Knots are a primary factor affecting the quality of sawn timber, making their detection fundamental to any timber grading or cutting optimization system. While X-ray computed tomography provides accurate knot locations and internal structures, it is often too slow or expensive for practical use. An attractive alternative is to use fast and cost-effective log surface measurements, such as laser scanners or RGB cameras, to detect surface knots and estimate the internal structure of wood. However, due to the small size of knots and noise caused by factors, such as bark and other natural variations, detection accuracy often remains low when only one measurement modality is used. In this paper, we demonstrate that by using a data fusion pipeline consisting of separate streams for RGB and point cloud data, combined by a late fusion module, higher knot detection accuracy can be achieved compared to using either modality alone. 
We further propose a simple yet efficient sawing angle optimization method that utilizes surface knot detections and cross-correlation to minimize the amount of unwanted arris knots, demonstrating its benefits over randomized sawing angles.
\end{abstract}










\end{frontmatter}



\section{Introduction}
\label{sec:introduction}
Defect detection is an essential part of quality control in almost all industrial processes. When dealing with natural materials, such as wood, the large visual and structural variation makes the defect detection task challenging. The task is further complicated in cases, where defects are small (e.g., knots in the wood) as labeling of large datasets for training is difficult. In practice, this means that the defect detection model should be able to locate subtle visual or structural cues that characterize defects and should be invariant to the large natural variation of materials. Due to these challenges, a single measurement modality, such as RGB images or laser point clouds, is often insufficient. However, by fusing information from multiple data modalities, the accuracy can be notably improved~\citep{Zhang_2023, chen20233d, zhao20223D}. 

The main defects defining the quality of timber in sawmill processes are knots~\citep{rais}, found at the locations where branches are attached to the tree. Depending on their location, the knots may compromise the structural integrity, appearance, and selling price of the end products resulting from the sawing process. This makes it important to detect the knots before sawing in order to allow process control (sawing optimization). Ideally, the internal structure of the knots would be directly measured using, for example, computed tomography (CT) scanners. However, such sensors are often expensive or slow, rendering them uneconomical for practical use in many sawmills. Recent studies~\citep{zolotarevKnotModelling, batrakhanov2021virtual} have demonstrated that internal knot distribution can be successfully estimated based on the surface locations of knots. This makes fast and inexpensive surface measurements an attractive alternative to CT scanners.

\begin{figure}
     \centering
     \begin{subfigure}[b]{0.4368\columnwidth}
         \centering
         \includegraphics[width=\columnwidth]{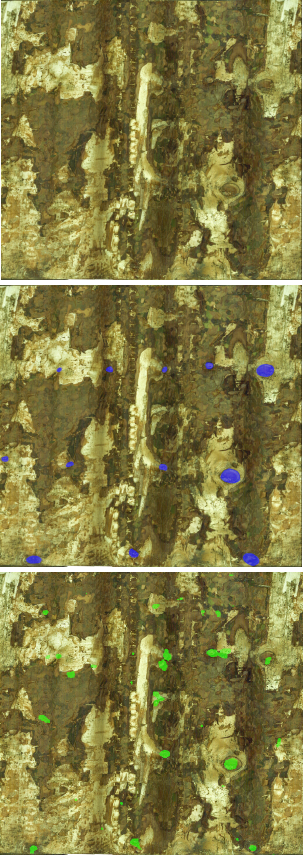}
         \caption{}
         \label{fig:taskRGB}
     \end{subfigure}
     \hspace{1em}
     \begin{subfigure}[b]{0.28\columnwidth}
         \centering
         \includegraphics[width=\columnwidth]{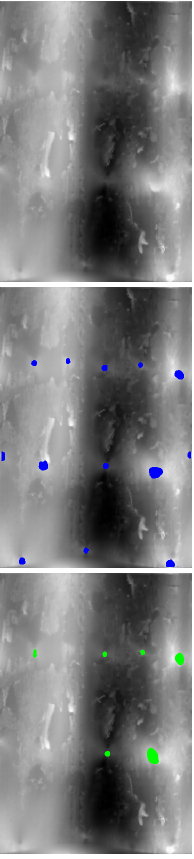}
         \caption{}
         \label{fig:taskHeight}
     \end{subfigure}
     \caption{Task illustration for image-based and point cloud-based methods. The top images show a sample of a log surface represented by an image (a) and a height map generated from point cloud (b). Middle images show the samples with X-ray-based annotations. The bottom row adds the model predictions based on only one modality to the samples.}
     \label{fig:taskChallenge}
\end{figure}

The two frequently used modalities to measure log surfaces are RGB images and laser-measurement-based point clouds. The small size of knots, combined with the large visual variation of bark and other surface characteristics, makes image-based detection of knots challenging even for humans (see Fig.~\ref{fig:taskRGB}). In point clouds, knots appear as small bumps on the surface, so distinguishing them from other surface irregularities is difficult, reducing the detection accuracy (see Fig.~\ref{fig:taskHeight}). Anyhow, neither modality alone tends to achieve high accuracy in knot detection.

\begin{figure*}
    \centering
    \includegraphics[width=0.72\linewidth]{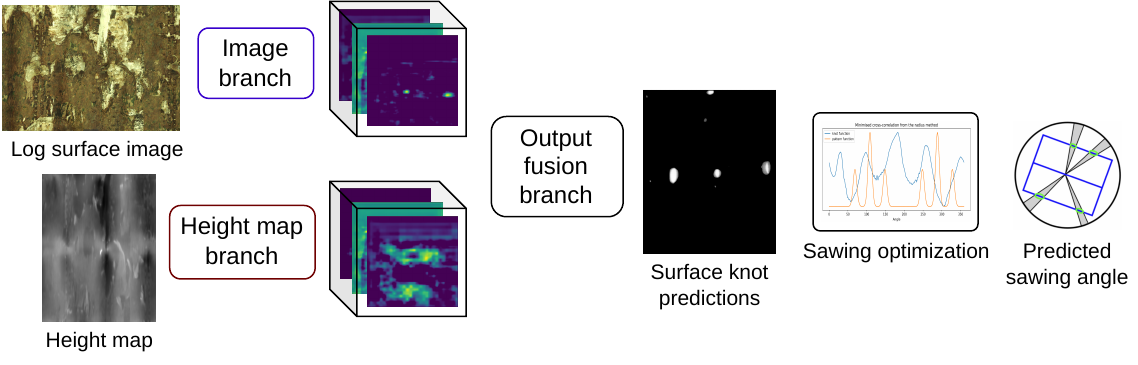}
    \caption{Overall fusion architecture diagram.}
    \label{fig:basicArch}
\end{figure*}

In this paper, we address this issue by proposing a data fusion pipeline for defect detection (see Fig.~\ref{fig:basicArch}). The proposed pipeline consists of separate RGB images and surface point clouds processing streams with fusion block combining the independent results. RGB images are processed with a Feature Pyramid Network-based (FPN) segmentation model~\citep{FPNSegmentation}. The point clouds are first converted into height maps and then processed with a separate FPN model. The benefits of separate streams are that (a) pretraining can be done separately for each stream, allowing the use of single-modality data in the training process; (b) intermediate results are available, making the method more transparent; (c) it is possible to align the modalities automatically by correlating the FPN outputs, and (d) detection results, albeit less accurate, can be obtained even if one of the modalities is missing. As the final step, the streams are combined in the output fusion module, which utilizes feature maps from the second-last layer of both FPN networks. The fusion module is composed of a set of convolution layers trained independently from the single-modality streams. It takes the concatenated feature maps as an input and produces the final knot detections. We further propose the process of obtaining aligned and accurately annotated training data by utilizing dense X-ray reconstructions and semi-automated point correspondence detection between the modalities.

\begin{figure}
    \centering
    \includegraphics[width=0.55\linewidth]{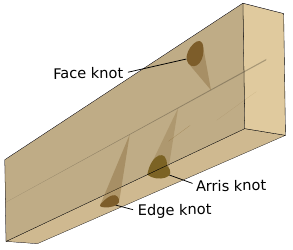}
    \caption{Illustration of different knot locations.}
    \label{fig:knotVariants}
\end{figure}

Finally, we extend the pipeline with a sawing optimization step. This is achieved with a simple yet efficient cross-correlation-based approach to estimate the sawing angle that minimizes the number of unwanted arris knots on the resulting boards. The arris knots together with other knot types are visualized in Fig.~\ref{fig:knotVariants}.

In the experimental part of this work, we demonstrate that, with proper pre-training, the proposed fusion pipeline is capable of learning with a very limited amount of training data and outperforms single modality-based detection models on challenging log data. We compare the proposed fusion strategy with the single-modality detectors and a fusion approach utilizing modality-specific predictions and show that it outperforms the other methods.

The main contributions of the paper are as follows: (a) a novel data fusion-based method for surface knot detection on wooden logs; (b) fast sawing angle estimation method for minimizing the number of unwanted arris knots; (c) analysis of the method on novel multi-modal log dataset.

\section{Related work}
\label{sec:relatedWork}

This section reviews relevant studies on timber sawing optimization, focusing on knot detection, segmentation techniques, various fusion approaches, and an evaluation of sawing optimization methods.

\subsection{Knot detection}
Knots are one of the main defects influencing the quality of wooden products. Consequently, the task of detecting knots is important; however, depending on the modality, it can be challenging as knots vary notably in size, shape, and appearance, especially with surface signs of knots being very subtle. 

Detection of internal knots is usually done from either images of sawn boards~\citep{ruzNeurofuzzy, urbonas2019Automated, todoroki2010Automated, zhang2023Threedimensional, norlander2015Wooden} or CT scans of logs~\citep{longuetaud2012Automatic,khazem2023Deep,giovanniniImproving, krahenbuhl2013Knot, krahenbuhl2014Knot}. Detecting knots from sawn boards is typically straightforward due to the high contrast between knots and the wooden board surface, allowing the use of simple approaches such as utilizing intensity differences~\citep{ruzNeurofuzzy, todoroki2010Automated, zhang2023Threedimensional} or using standard CNN-based methods~\citep{norlander2015Wooden,urbonas2019Automated}. These proved to be of very high accuracies but for process optimization purposes, it is important to detect the knots from logs before sawing.
One option to do this is to utilize CT scans. Knots have higher density than the rest of the wood; therefore, they can be segmented from X-ray data using thresholding and shape information~\citep{longuetaud2012Automatic, krahenbuhl2013Knot, krahenbuhl2014Knot} or by using CNN~\citep{khazem2023Deep,giovanniniImproving}.

Segmentation of surface knots is usually performed on point clouds of tree trunks or wooden logs. The prevailing approach for surface knot segmentation is to utilize a transformation into a cylindrical or log-centric coordinate system and performing segmentation using hand-crafted features \citep{thomas2007Robust, thomas2011Graphical, thomas2007Automated, kretschmer2013New, zolotarevKnotModelling, van-thonguyen2016Segmentation}. To the best of our knowledge, no methods for surface knot segmentation on point clouds using deep learning methods are known. 

\subsection{Image and point clouds fusion}
Point cloud segmentation is a fundamental task in 3D computer vision, driving applications such as autonomous driving, robotics, and AR or VR. However, deep learning methods for point clouds must address inherent challenges such as irregularity, sparsity, and lack of grid structure issues not encountered in conventional image data processing.

\subsubsection*{Point cloud segmentation}
PointNet is a pioneering method that directly processes unordered point sets, utilizing multi-layer perceptrons (MLPs) to extract point-wise features and a global aggregation mechanism \citep{qi2017PointNet}. Building on this, PointNet++ introduces a hierarchical structure that captures local geometric features at multiple scales, enhancing performance on complex point cloud data \citep{qi2017PointNeta}. Point Transformer integrates self-attention mechanisms to model relationships between points, thereby improving segmentation accuracy in complex scenes by focusing on contextual similarities \citep{zhao2020Point}. Point Transformer v2 further refines the attention mechanism for greater efficiency and accuracy \citep{wu2022Point}, while Point Transformer v3 incorporates hierarchical attention and adaptive down-sampling, enabling better handling of high-resolution point clouds \citep{wu2023Point}.

Other approaches, such as OctNet \cite{riegler2017OctNet}, use octree-based structures to efficiently process only the occupied regions of 3D space, reducing memory and computational requirements and making them suitable for high-resolution point clouds. PAConv introduces adaptive convolution kernels that dynamically adjust to local point structures, capturing complex geometries for more effective segmentation \citep{xu2021PAConv}. PolarNet transforms point clouds into a polar coordinate system, offering a simplified representation ideal for large-scale outdoor scenes, like those encountered in autonomous driving, and achieving both efficient and accurate segmentation \citep{zhang2020PolarNet}. DGCNN constructs dynamic graphs from point clouds to capture spatial and feature-based relationships, using edge convolutions to achieve state-of-the-art segmentation performance \citep{wang2019Dynamic}.

\subsubsection*{Point cloud fusion with other modalities}
The fusion of point clouds with other data modalities, such as RGB images, further enhances the robustness of point cloud-only methods. These multi-modal approaches integrate visual and geometric data to improve accuracy in tasks like object detection, segmentation, and alignment.

While point clouds hold accurate geometric information, by utilizing a complementary modality it is possible to achieve better accuracy or accomplish more complicated tasks. Many different fusion strategies exist, but, generally, the two main approaches are (a) to combine modalities and process them together (add information from an other modality to point cloud such as corresponding pixel value), or (b) process them independently and share information (two pipelines connected in various steps and sharing extracted features). For instance, \cite{chen20233d} combine RGB data with point clouds to improve object differentiation and spatial context, while \cite{zhao20223D} fuse high-level features from both modalities to enhance 3D vehicle detection, particularly in occluded scenarios. Similarly, \cite{ma2019Accurate} integrate color information into 3D reconstructions to enhance monocular 3D object detection by leveraging both geometric and visual cues.

In point cloud alignment, \cite{yuan2023PointMBF} propose a multi-scale bidirectional fusion network for unsupervised RGB-D point cloud alignment, effectively combining multiple modalities to improve alignment accuracy. Likewise, \cite{madawy2019RGB} merge dense LiDAR spatial data with RGB semantic content, enhancing segmentation, especially for smaller objects, demonstrating the benefit of combining depth and color data. Finally, \cite{salazar-gomez2022TransFuseGrid} use transformers to capture long-range dependencies between LiDAR and RGB data, which improves semantic grid predictions in complex environments.

\subsection{Sawing optimization}

Multiple studies highlight the advantages of sawing optimization methods that integrate CT scanning and virtual simulation to maximize lumber yield and value. \cite{chang} used medical X-ray CT scanning and the TOPSAW sawing optimization software to demonstrate a $46\%$ higher value yield compared to live sawing, with notable gains in low-grade logs. \cite{lundahl} optimized rotation and positioning using virtual sawing simulations and observed a value increase of up to 22\% and a volume yield improvement of $4.5\%$, emphasizing the sensitivity of yield to minor positioning errors.

\cite{rais} employed Monte Carlo simulation to optimize sawing angles based on internal knot distributions, reporting a relative value increase of $4\text{-}20\%$ across different price scenarios. Similarly, \cite{stangle} combined CT-scanned defect data with cutting simulations, achieving a $24\%$ increase in volume and a $13\text{–}24\%$ gain in value yield, particularly for low-grade logs. \cite{fredriksson} conducted an extensive study involving 712 pine and 750 spruce logs, showing a 21\% value gain when optimizing based on internal and external log characteristics, with a $13\%$ increase using only external data.

\section{Proposed method}
\label{sec:proposedMethod}

The proposed method for surface knot detection utilizes image and surface point cloud data. Prior to detection, the surface RGB image of the whole log is stitched from individual images captured from different angles to form an image of the log's surface. The point cloud data is converted to a height map image and aligned to the coordinate system of the RGB data. Subsequently, the knots are detected from the stitched RGB images and height maps using independent processing branches to obtain modality-specific feature maps. Feature maps from both branches are fused to produce the final knot detections. As it is not possible to rule out that only one modality is available, the model can be also used with only one modality. Finally, the knot detections are used to search for the optimal sawing angle. Here, the optimization is defined as minimizing the number or area of arris knots in the final boards. The full pipeline is shown in Fig.~\ref{fig:overallPipeline}.

\subsection{Height map computation}
\label{sec:heightmap}
The point clouds are converted to height maps using the method described in~\cite{zolotarevKnotModelling}. The conversion from point cloud to height map is performed by Cartesian to log-centric coordinates. As logs can be curved, we first divide the log's center line into a chosen number of segments and separate the point cloud with planes bisecting an angle between the center line segments. The location of each point is converted into the cylindrical coordinate system in regard to a given segment acting as the $z$-axis, resulting in $\theta, \rho$ and local $z$ coordinates. 

The height map is then defined as the values of function $f$ on an evenly spaced 2D grid of values $\theta$ and $l$. The coordinate $l$ is computed by adding sums of lengths of previous center lines. The heightmap is found by fitting $f$ to the point cloud. As $f$ is assumed to be smooth, we apply regularization to the gradient, and it is so possible to control the smoothness of the generated height map. A more in depth description of the method can be found in \citep{zolotarevKnotModelling}.

\subsection{RGB image stitching}
\label{sec:stitching}
\begin{figure}[ht]
    \centering
    \includegraphics[width=0.95\linewidth]{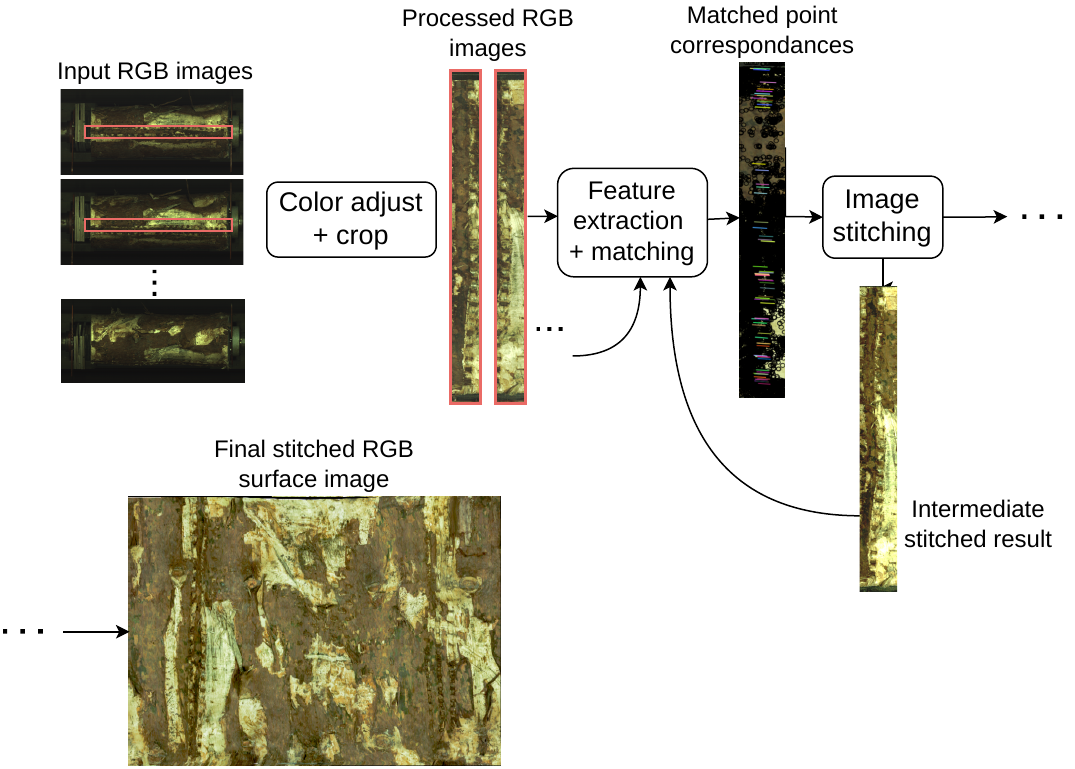}
    \caption{RGB image stitching diagram. Images are first cropped, adjusted, and then iteratively stitched forming the final stitched RGB surface image.}
    \label{fig:imgStitch}
\end{figure}

For RGB image stitching, the original images are first cropped and color-corrected. For this purpose, surface RGB images containing only the region of interest (log's surface) are obtained. In our case, we perform image cropping to mitigate the influence of lens distortion and acquire only the region of interest (surface of a log). Images are subsequently converted to cylindrical coordinates, and SIFT feature points are extracted and matched between the images~\citep{lowe2004DistinctiveIF}. A perspective transformation is estimated, and finally, the images are stitched. The process is performed recursively, adding a new image "stripe" to the stitched image at each step, as illustrated by Fig.~\ref{fig:imgStitch}.

\subsection{Single modality knot detection}
\label{sec:singleModalBranches}
\begin{figure*}[t]
  \centering
  \includegraphics[width=\textwidth]{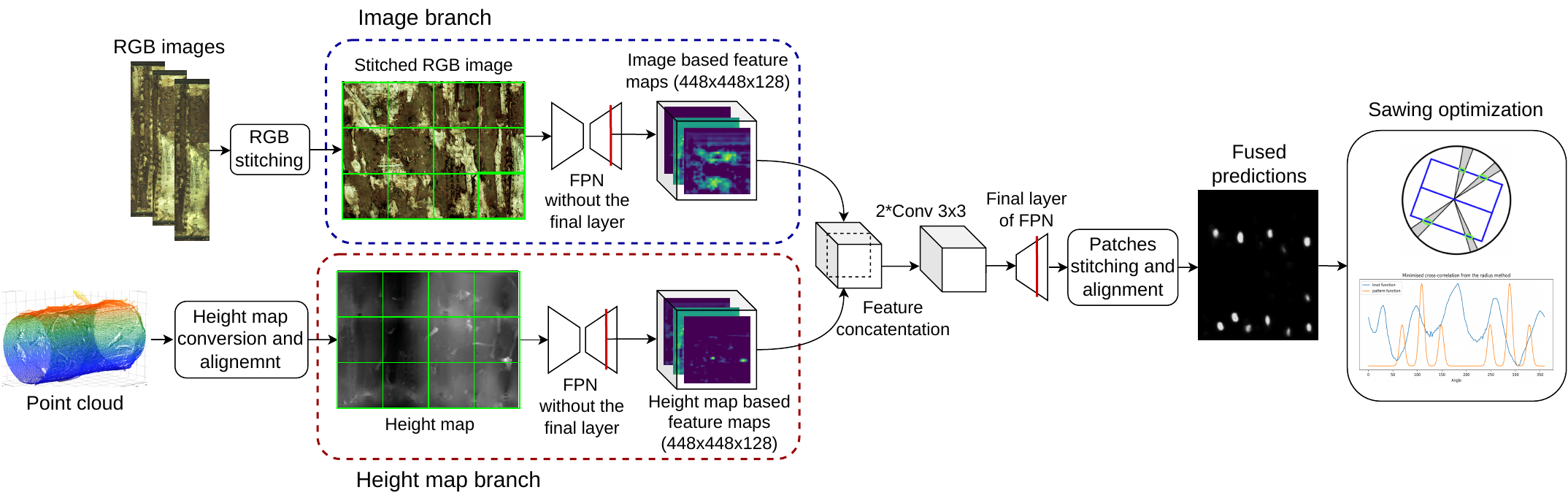}
  \caption{Overall sawing optimization method pipeline.}
  \label{fig:overallPipeline}
\end{figure*}

The proposed method can be divided into a height map branch, an image branch, and a fusion output block. The architecture of the RGB image processing and the height map processing branch is identical except for the number of input channels: one for the height map branch and three for the RGB branch. Both of them can also be used independently if only one modality is available. The single modality branches use the FPN-based encoder-decoder segmentation model~\cite{FPNSegmentation} with MobileNet V3~\cite{MobileNetv3} encoder to keep the number of parameters low and to make the model less likely to overfit on small data sets. In order to preserve the fine features present in the RGB data during the detection step, both modalities are converted to patches. In this way, the feature maps of both modalities can be fused together. 
While both branches produce probability maps and so can be used independently, in our fusion method, we use feature maps from the last layer of the FPN networks (more in Subsection~\ref{sec:dataFusion}). Fig.~\ref{fig:overallPipeline} shows the overall architecture with the image and height map branches highlighted.

\subsection{Data fusion}
\label{sec:dataFusion}

For data fusion, the feature maps from the last layer of the modality-specific FPN networks are utilized. Each feature map contains 128 channels as it follows the FPN layer structure. These feature maps are concatenated into a single feature representation, which is then fed into a fusion network. This network consists of three convolutional layers with ReLU activation, producing a prediction mask for each patch. Finally, the patches are stitched together to generate the final prediction mask. By fusing feature maps instead of modality-specific predictions allows to transfer more information from the data to the fusion module.

\subsection{Sawing optimization}
\label{sec:sawOptimMethod}
The main factor affecting the quality and the grade of the sawn timber is the types and positions of knots. Ideally, the knots should be on either of the faces, while knots on the edges of the board, especially arris knots (intersection of a face and an edge), should be avoided as they compromise the structural integrity. We consider the sawing optimization from the point of view of minimizing the number of arris knots. The optimization is performed by searching for the optimal sawing angle (rotation angle around the centerline of the log) that is estimated to minimize the number of arris knots by utilizing the surface knot detections from the data fusion step. To perform the optimization efficiently, we reduce the problem to 2D by only considering knot and corner angles around the centerline and ignoring the knot locations in longitudinal direction. This way, the knot angles and and the sawing pattern can be presented as 1D functions and the optimization can be formulated as searching of shift that minimizes the correlation between the functions.   
The proposed method consists of three steps:
\begin{enumerate}
    \item construction of a pattern function,
    \item construction of a knot function, and
    \item minimization of cross-correlation between both functions.
\end{enumerate}

\subsubsection*{Pattern function}
\begin{figure}[ht]
    \centering
    \begin{subfigure}[b]{0.45\columnwidth}
        \centering
        \includegraphics[scale=0.15]{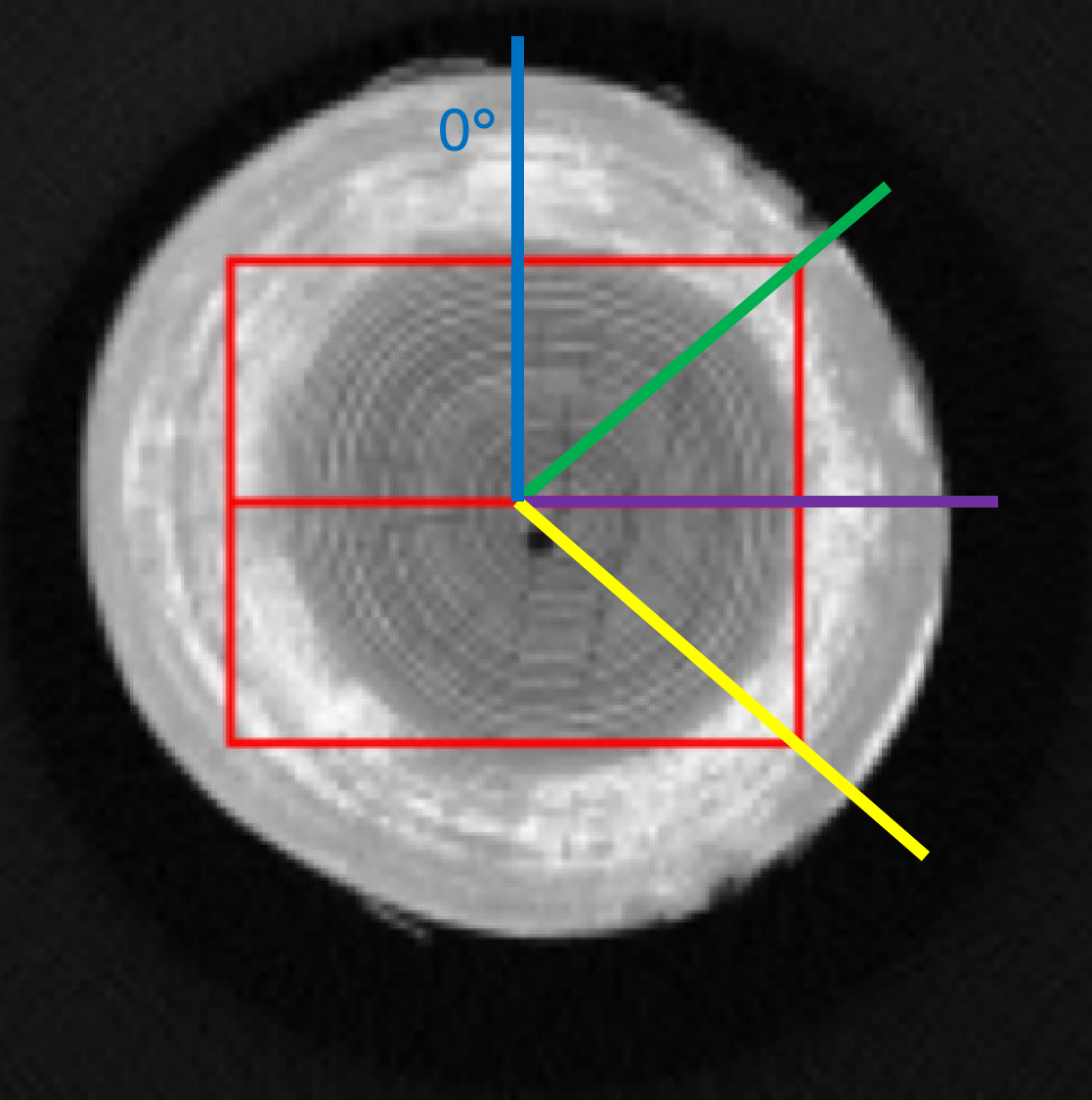}
        \caption{Sawing pattern.}
    \end{subfigure}
    \hfill
    \begin{subfigure}[b]{0.45\columnwidth}
        \centering
        \includegraphics[width=\columnwidth]{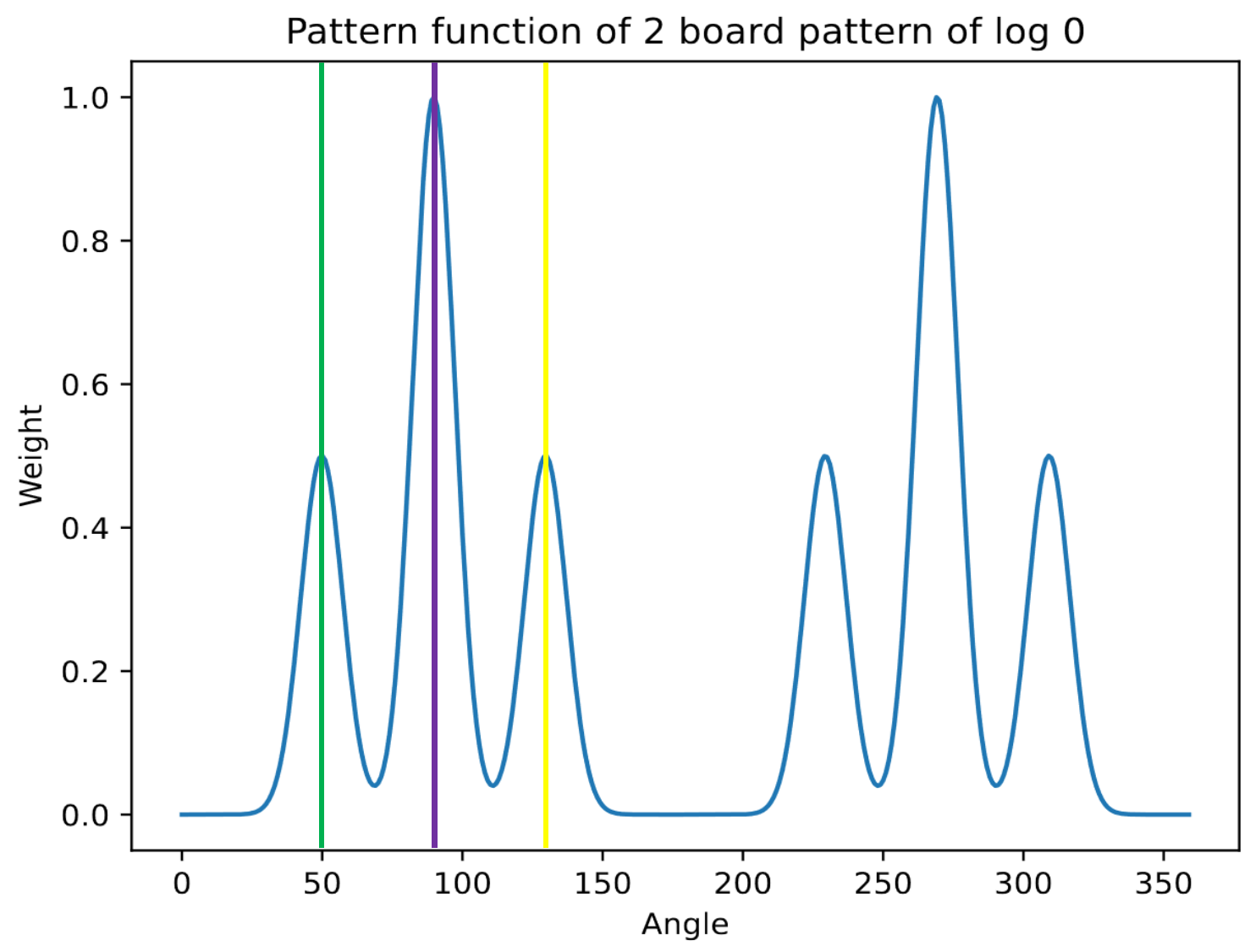}
        \caption{Corresponding pattern function.}
    \end{subfigure}
    \caption{(a) An example of a two-board sawing pattern. (b) The pattern function for this pattern. The colored lines correspond to each other.}
    \label{fig:pattern-function}
\end{figure}

The pattern function describes the sawing pattern, especially the angles corresponding the corners of planks or boards. The first step is to locate the corner points of the boards in the sawing pattern and to calculate their polar angles. To address the uncertainty and to obtain continuous function, corners are modeled as Gaussians. This means the pattern function~$f_p$ is defined as mixture of Gaussians:
\begin{equation}
\label{eq:pattern-function}
    f_p(\theta) = \sum_{\theta \in S} \mathcal{N}(\theta,\,\sigma^{2}),
\end{equation}
where $\theta$ represents the polar angle, $\theta \in \, \left[0, \, 360\right), \ \theta \in \mathbb{R}$, $S$~is a set of polar angles of the pattern's corner points, and $\sigma$ is estimated to be the same as the standard deviation of a real knot. An example of it is depicted in Fig.~\ref{fig:pattern-function}.

\subsubsection*{Knot function}
\begin{figure}[ht]
    \centering
    \begin{subfigure}[b]{0.32\columnwidth}
        \centering
        \includegraphics[width=\columnwidth]{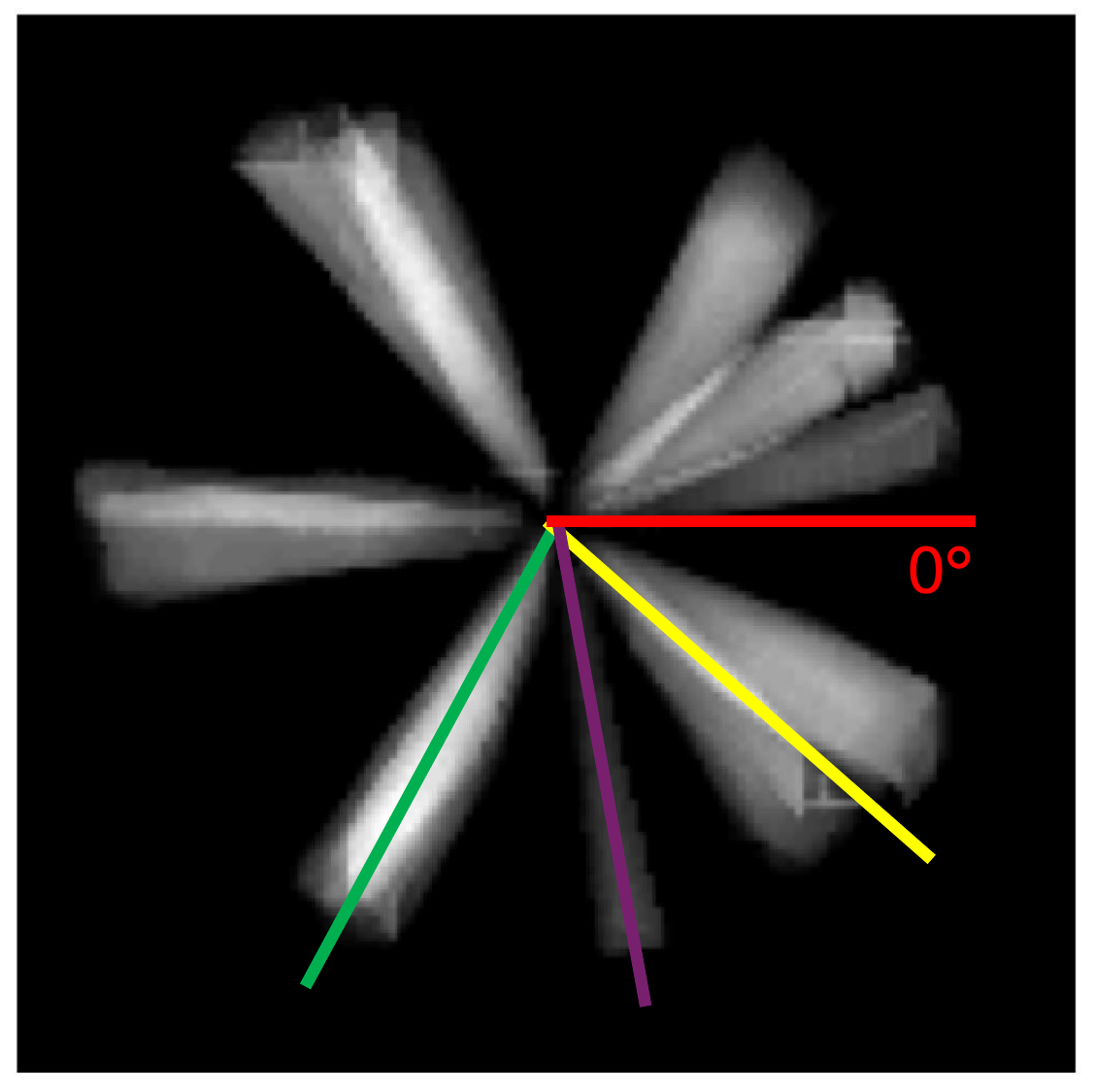}
        \caption{}
    \end{subfigure}
    \hfill
    \begin{subfigure}[b]{0.32\columnwidth}
        \centering
        \includegraphics[scale=0.15]{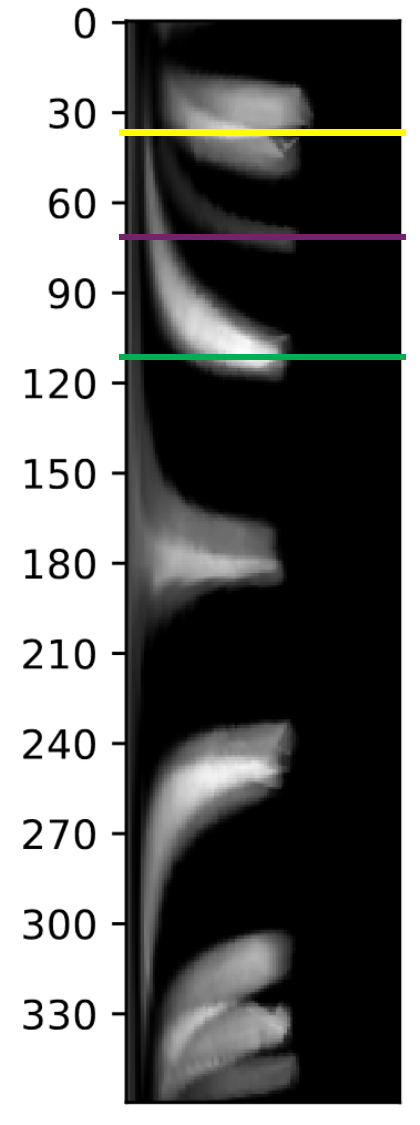}
        \caption{}
    \end{subfigure}
    \hfill
    \begin{subfigure}[b]{0.32\columnwidth}
        \centering
        \includegraphics[width=\columnwidth]{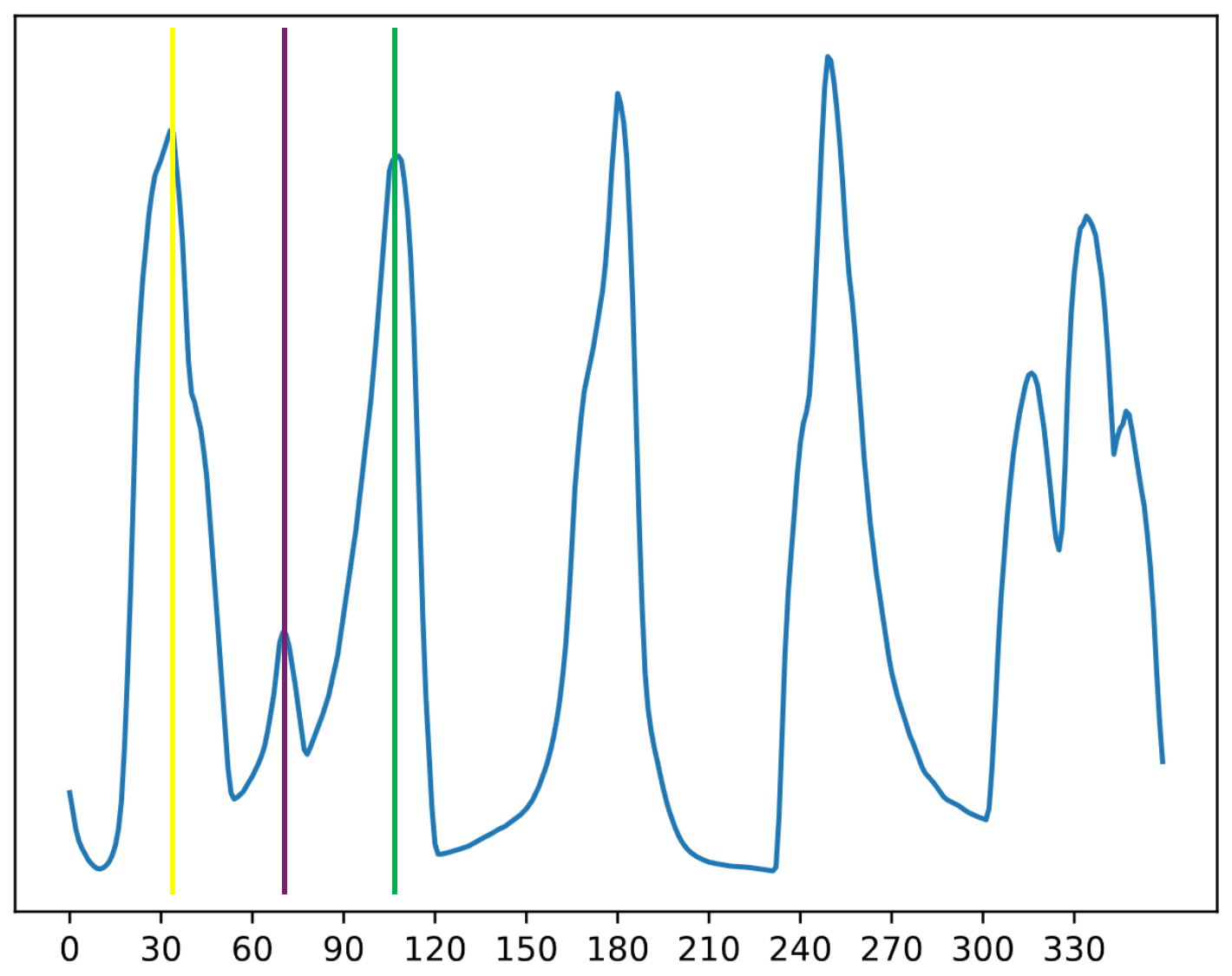}
        \caption{}
    \end{subfigure}
    \caption{(a) A projection of detected knots. (b) The polar representation of the projection. (c) The extracted knot function. The colored lines correspond to each other.}
    \label{fig:knot-function}
\end{figure}

The knot function describes the angles of found knots. In practice, each angle is assigned a value that corresponds to the probability of finding a knot at a specific polar angle. The process is shown in Fig.~\ref{fig:knot-function}. First the detected knots are projected to 2D cross-section by assuming that the knots grow directly from centerline to surface, consequently the projection is converted to polar coordinates. Finally, knot function is calculated from the polar representation. Mathematically, the knot function $f_k$ is defined as
\begin{equation}
\label{eq:knot-function}
    f_k : \theta \rightarrow k,
\end{equation}
where $\theta \in \, \left[0, \, 360\right)$ represents the polar angle, and $k$ represents the extracted knot appearance. The knot function can be acquired by summing and normalizing the prediction mask along the y-axis.

\begin{figure}[ht]
    \centering
    \includegraphics[width=0.3\linewidth]{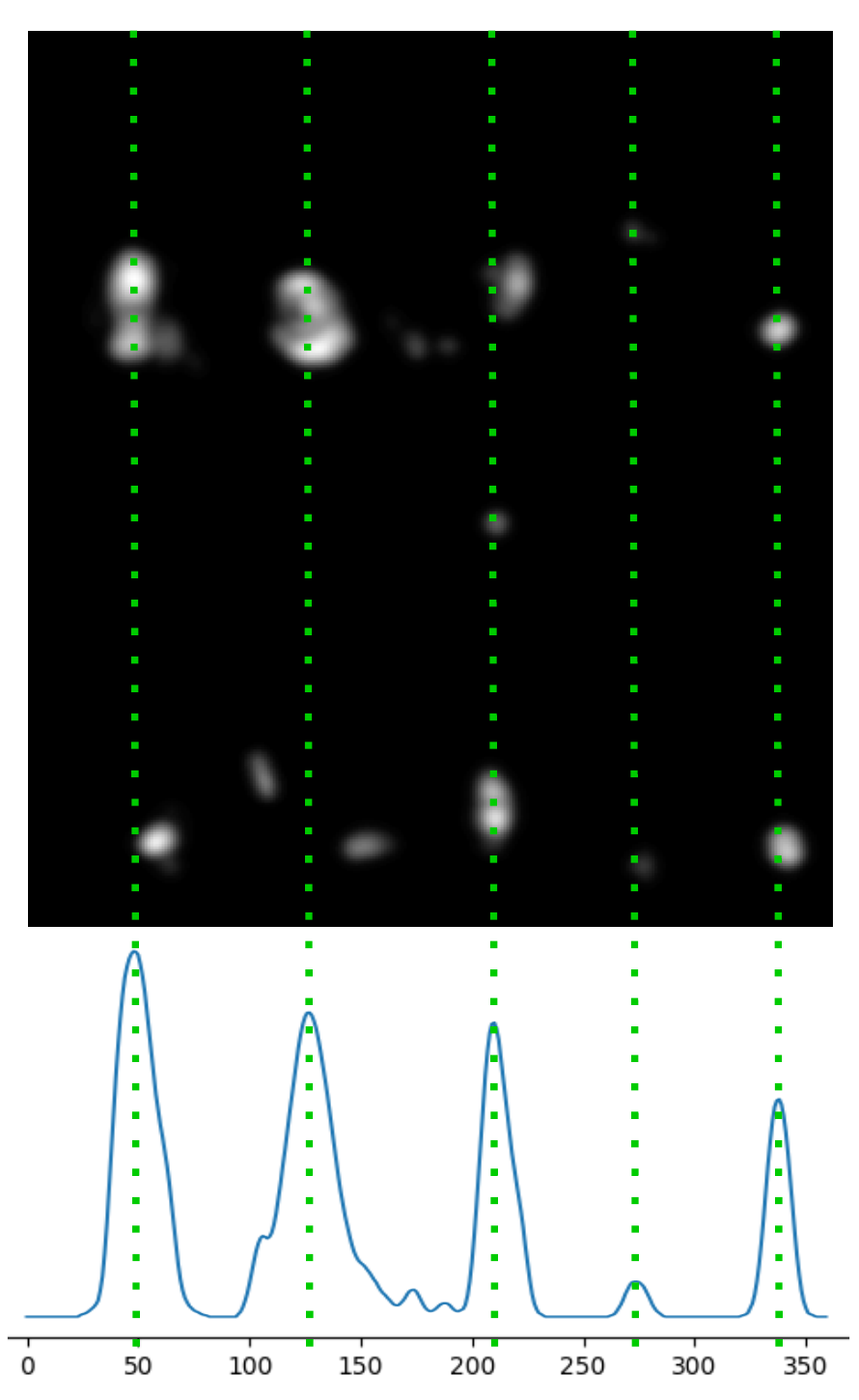}
    \caption{Height map to knot function conversion illustration.}
    \label{fig:conversion}
\end{figure}

\subsubsection*{Minimization of cross-correlation}
The last step of the angular optimization method is to find the offset from the knot and pattern functions. This offset will represent the angle of rotation for sawing. Cross-correlation, in this case, computes how similar the computed functions are. Minimizing its result means that the knots will avoid corners of boards, thus minimizing the arris knot appearances. In case of sawing patterns that are two-fold rotational symmetric, the process can be reduced to a lower range of angles.

\section{Results}
\label{sec:experiments}

\subsection{Data}
\label{sec:data}
\begin{figure}[ht!]
     \centering
     \begin{subfigure}[b]{0.24\columnwidth}
         \centering
         \includegraphics[width=\columnwidth]{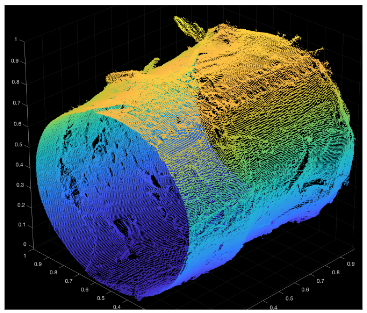}
         \caption{}
         \label{fig:laserExample}
     \end{subfigure}
     \hfill
     \begin{subfigure}[b]{0.36\columnwidth}
         \centering
         \includegraphics[width=\columnwidth]{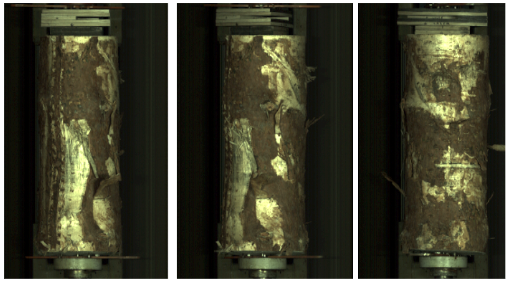}
         \caption{}
         \label{fig:rgbExample}
     \end{subfigure}
     \hfill
     \begin{subfigure}[b]{0.31\columnwidth}
         \centering
         \includegraphics[width=\columnwidth]{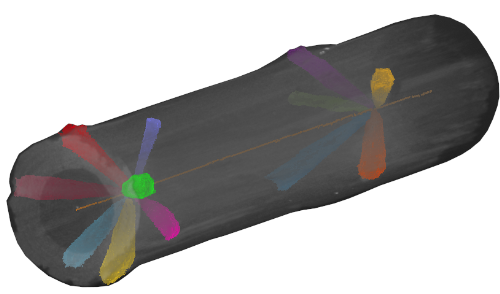}
         \caption{}
         \label{fig:ctExample}
     \end{subfigure}
     \caption{Data samples for each modality (a) laser point cloud (b) RGB images (c) CT-scan reconstruction with manual annotations.}
     \label{fig:example}
\end{figure}

The main dataset consists of X-ray CT reconstructions, surface laser point clouds, and RGB images of nine Scots pine logs without debarking acquired in a laboratory environment. The X-ray CT reconstructions were created using dense reconstruction, and the resulting volumetric data have a resolution of $128\times 128\times x$ voxels, where log lengths determine $x$. The image data consists of $2592\times1593$ pixel images captured with a stationary camera and a rotating log with one image per degree of rotation. The surface laser point clouds were obtained by scanning debarked log surfaces before sawing. The minimum distance between two neighboring points is 0.05 mm. Fig.~\ref{fig:example} shows examples from each modality. The ground truth (GT) data was obtained by manually annotating the CT reconstructions where the knots are clearly visible, having a higher density than the rest of the wood, and finding locations where the knots reach the log surface.

In addition to this, 50 debarked Scots pine logs with only laser point cloud data were used for pretraining purposes. Due to the debarking,  the surface knots were more visible, and therefore, reasonably accurate GT for knot locations was possible to obtain using earlier automatic method~\citep{zolotarev2020Modelling}. 

\subsection{Data preprocessing}
Various preprocessing steps were applied to allow method for training and evaluation. The ground truth information, present only in the CT reconstructions, needed to be transferred to other modalities (see Fig.~\ref{fig:example}). This was done in three stages. First, point clouds were aligned with CT reconstructions to obtain the GT knot locations for point clouds. Then, these locations were transferred to the height maps. Finally, the height maps were aligned with the stitched RGB image to obtain the GT for the RGB images.

\subsubsection*{CT data to laser point clouds mapping}
\begin{figure}[ht]
    \centering
    \includegraphics[width=0.9\linewidth]{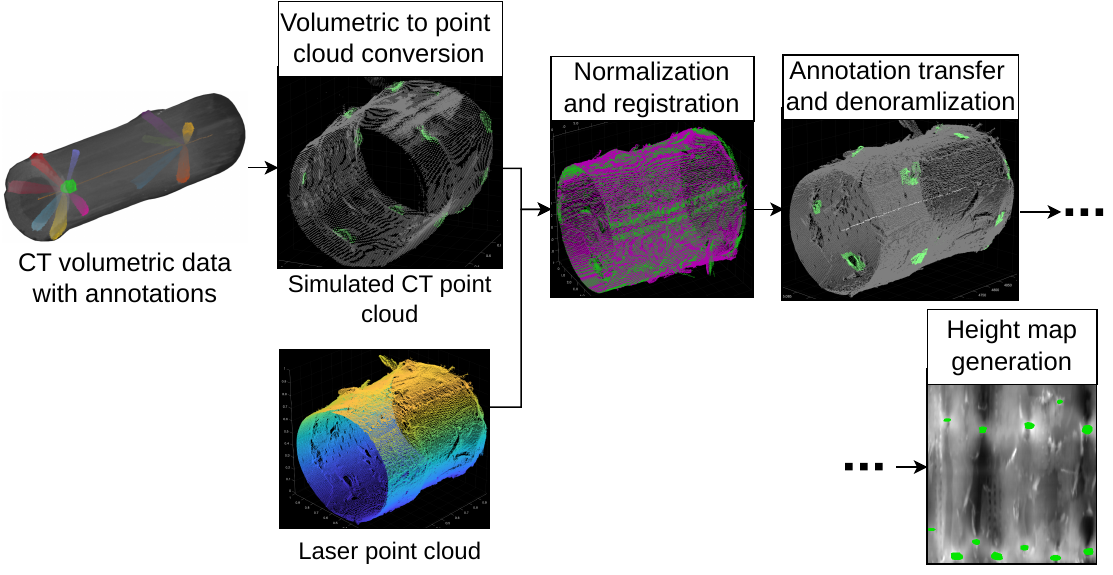}
    \caption{Modality alignment pipeline.}
    \label{fig:GTGeneration}
\end{figure}

In order to transfer the GT annotations to point clouds, simulated surface point clouds were generated from the CT scans based on known density information. These simulated point clouds were registered (aligned) to the real laser point clouds. This was done first by normalizing point clouds by calculating centroids, shifting to the coordinate origins, and scaling based on the furthest distant points. We then applied the iterative closest point algorithm~\citep{ICP}. The GT information was transferred to the laser point clouds based on point-wise distance and reverted to the original coordinate system. Laser point clouds with annotations were subsequently converted to the height map representation as described in the method section. Fig.~\ref{fig:GTGeneration} illustrates the alignment process. 

\subsubsection*{Image data to laser point clouds mapping}
\begin{figure}[ht]
    \centering
    \includegraphics[width=1\linewidth]{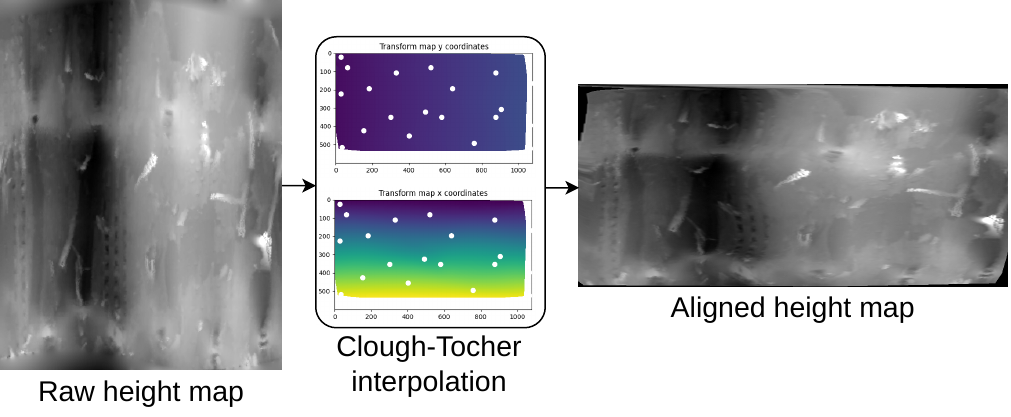}
    \caption{Height map alignment based on point correspondences using Clough-Tocher interpolation.}
    \label{fig:ImageHeightAlign}
\end{figure}

The next step was to align the image data with the height maps to transfer the annotations to the RGB data. To cover the whole log surface and to obtain a similar representation to height maps, the individual images were stitched as described in Sec.~\ref{sec:stitching}.
To align the stitched image of the surface with the height map, we first manually identified corresponding feature points. Using the Clough-Tocher interpolation~\citep{cloughTocher} and the point correspondences as input, and so mappings between the images and height maps were calculated. Finally, the GT annotations were transferred from height maps to the RGB images (see Fig.~\ref{fig:ImageHeightAlign}).

\subsection{Model training}
For training, we used cross validation approach so every sample was used as the test partition and the remaining samples for validation and training, creating multiple models, one for each log. We used standard augmentation methods, such as affine transformations for both modalities and color jitter and elastic transformations for the RGB image data.

As the number of samples that included all the modalities was limited, we had to independently pre-train our models for image data and models for height map data. We used the additional dataset of 50 debarked Scots pine logs with only laser point cloud data and no ground truth annotations. To utilize these measurements, we used the knot segmentation method proposed by \cite{zolotarevKnotModelling}, where knots are segmented using Laplacian of Gaussian filter. These weak annotations were then used to pre-train the height map branch. For the image branch, we utilized encoders pretrained on the ImageNet dataset~\citep{imagenet}.

For the training of single-modality branches, we used a Binary Cross Entropy loss function and an adaptive RMSprop optimizer for height map model pre-training and image model training. For height map model fine-tuning, we then used SGD with a small learning rate. The size of FPN inputs was $448\,\times\,448\,\times\,1$ for the height map model and $448\,\times\,448\,\times\,3$ for the RGB model. For pre-training and fine-tuning, we utilized batch size 8 for every model. The fine-tuning learning rate for the SGD optimizer was $10^{-6}$ with a learning rate decay of $10^{-8}$.

For the training of the fusion branch, we also used a Binary Cross Entropy loss function and an adaptive RMSprop optimizer. As input, we used concatenated feature maps of size $448\,\times\,448\,\times\,128$ to form feature maps with size $448\,\times\,448\,\times\,256$, as describer in Section~\ref{sec:dataFusion}.

\subsection{Knot detection}
\label{sec:defDet}

While we use segmentation models for defect detection, the knots do not typically have well-defined borders on surface, and the annotations are, in some cases, unambiguous. While the segmentation evaluation metrics such as dice-Sørensen coefficient and Intersection over Union (IoU) are useful for training, they are not ideal metrics for evaluating the methods' performance. 
It is better to utlize detection metrics for the task we are more interested whether all knots are correctly detected than the shape of the segmenetation masks. The metric we use is the mean average precision (mAP)~\citep{NIPS2015_14bfa6bb} as that is the most commonly used for evaluating detection accuracy. We used relatively small IoU threshold (10\%) for true positives.

Table~\ref{tab:detMet} shows mAP results for all nine logs. We compared the proposed data fusion-based method to single-modality-based knot detection methods and a fusion approach utilizing single-modality predictions. The single-modality-based methods correspond to the individual branches of the proposed pipeline. The baseline fusion method (max) detects the knots by computing pixel-wise maximum value of the two knot probability maps from single-modality branches. As can be seen the proposed fusion approach clearly outperforms the other methods.

\begin{figure}[ht!]
     \centering
     \begin{subfigure}[b]{0.22\columnwidth}
         \centering
         \includegraphics[width=\columnwidth]{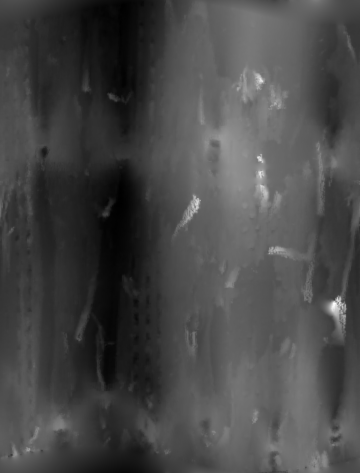}
         \caption{}
         \label{fig:rawHeight1}
     \end{subfigure}
     \hfill
     \begin{subfigure}[b]{0.35\columnwidth}
         \centering
         \includegraphics[width=\columnwidth]{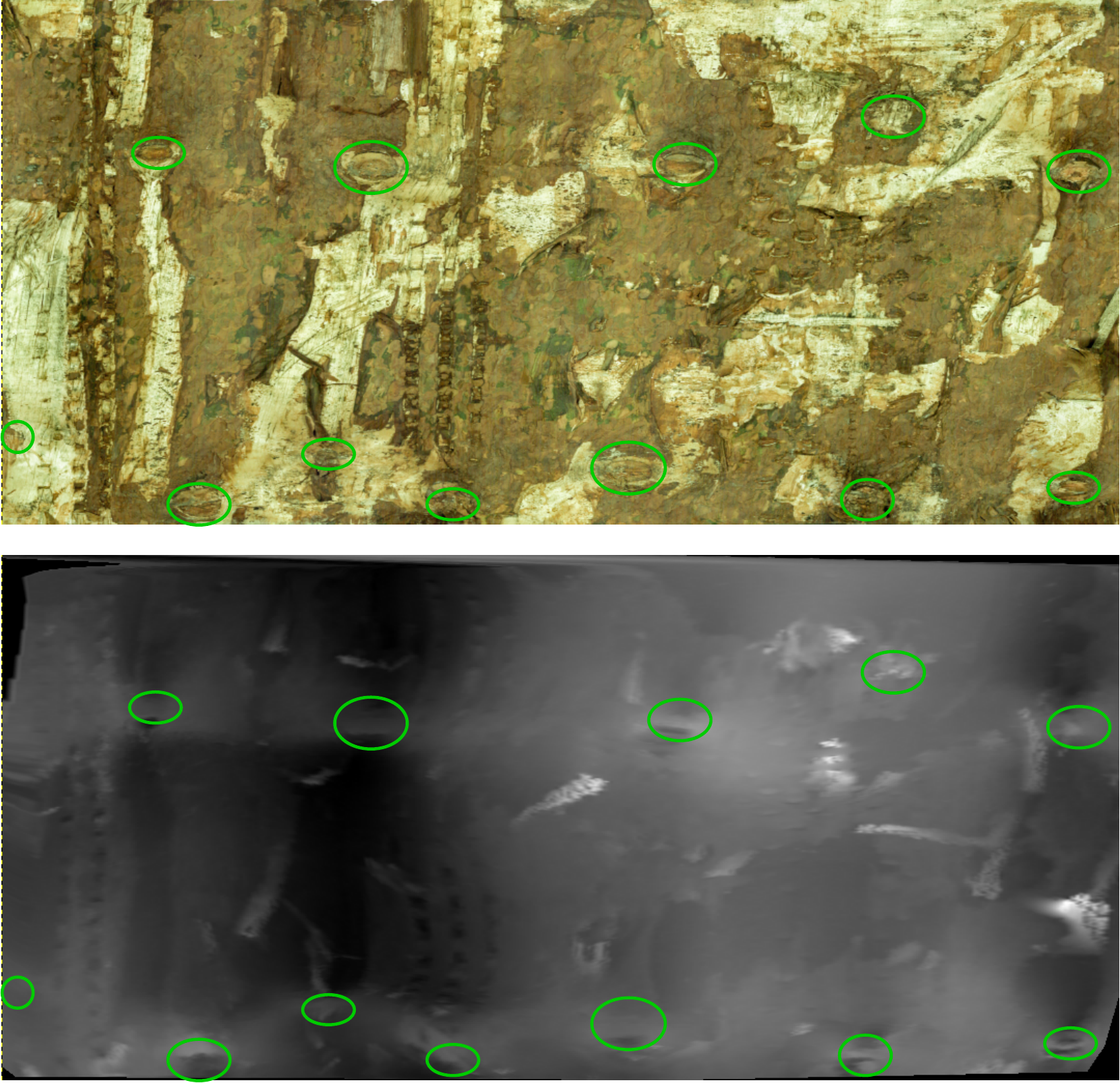}
         \caption{}
         \label{fig:modalsAligned1}
     \end{subfigure}
     \hfill
     \begin{subfigure}[b]{0.3\columnwidth}
         \centering
         \includegraphics[width=\columnwidth]{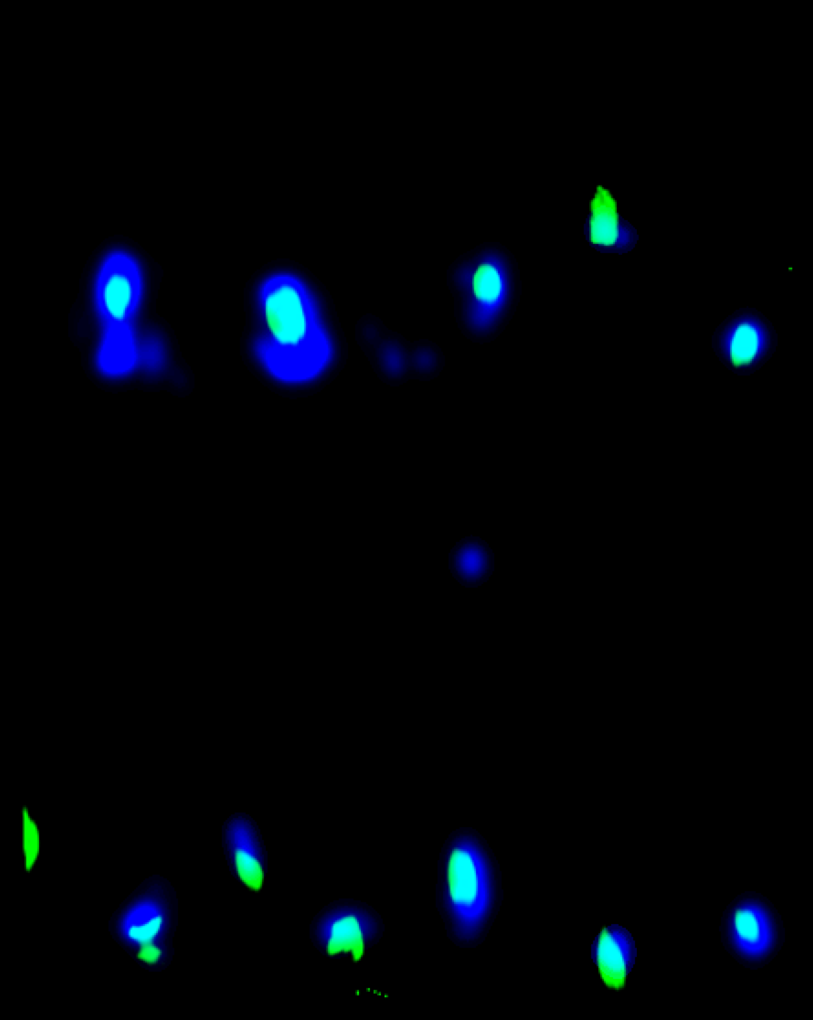}
         \caption{}
         \label{fig:resultsDiff1}
     \end{subfigure}
     \caption{Prediction example of log 1 compared to ground truth (a) raw height map (b) aligned RGB image and height map (c) predictions compared to ground truth (blue -- predictions, green -- ground truth, cyan -- correct predictions).}
     \label{fig:predExample1}
\end{figure}

\begin{figure}[ht!]
     \centering
     \begin{subfigure}[b]{0.22\columnwidth}
         \centering
         \includegraphics[width=\columnwidth]{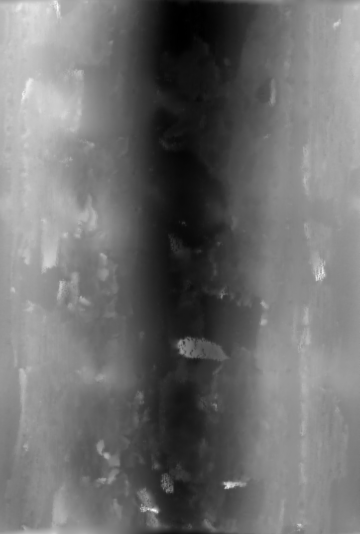}
         \caption{}
         \label{fig:rawHeight8}
     \end{subfigure}
     \hfill
     \begin{subfigure}[b]{0.24\columnwidth}
         \centering
         \includegraphics[width=\columnwidth]{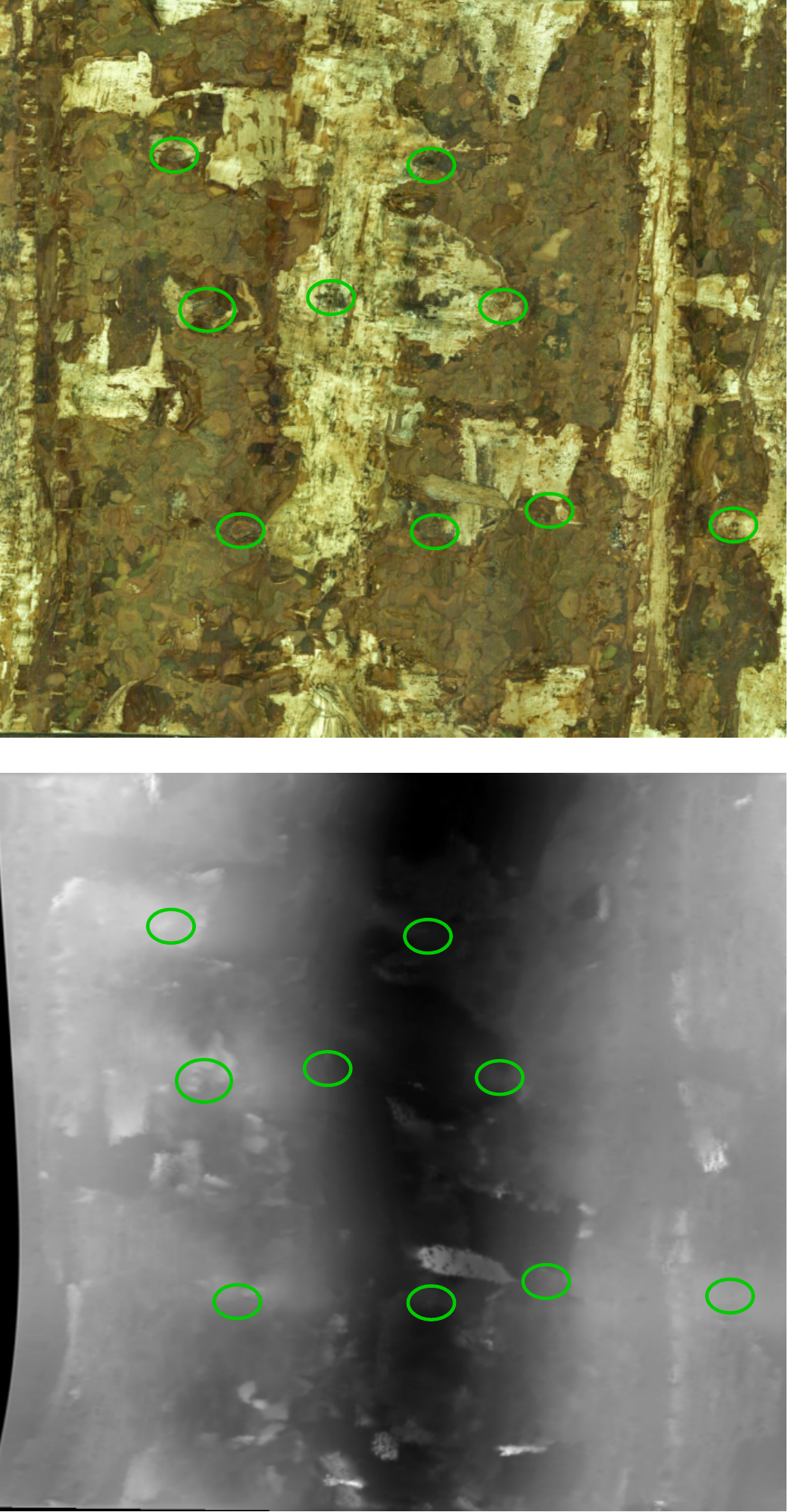}
         \caption{}
         \label{fig:modalsAligned8}
     \end{subfigure}
     \hfill
     \begin{subfigure}[b]{0.33\columnwidth}
         \centering
         \includegraphics[width=\columnwidth]{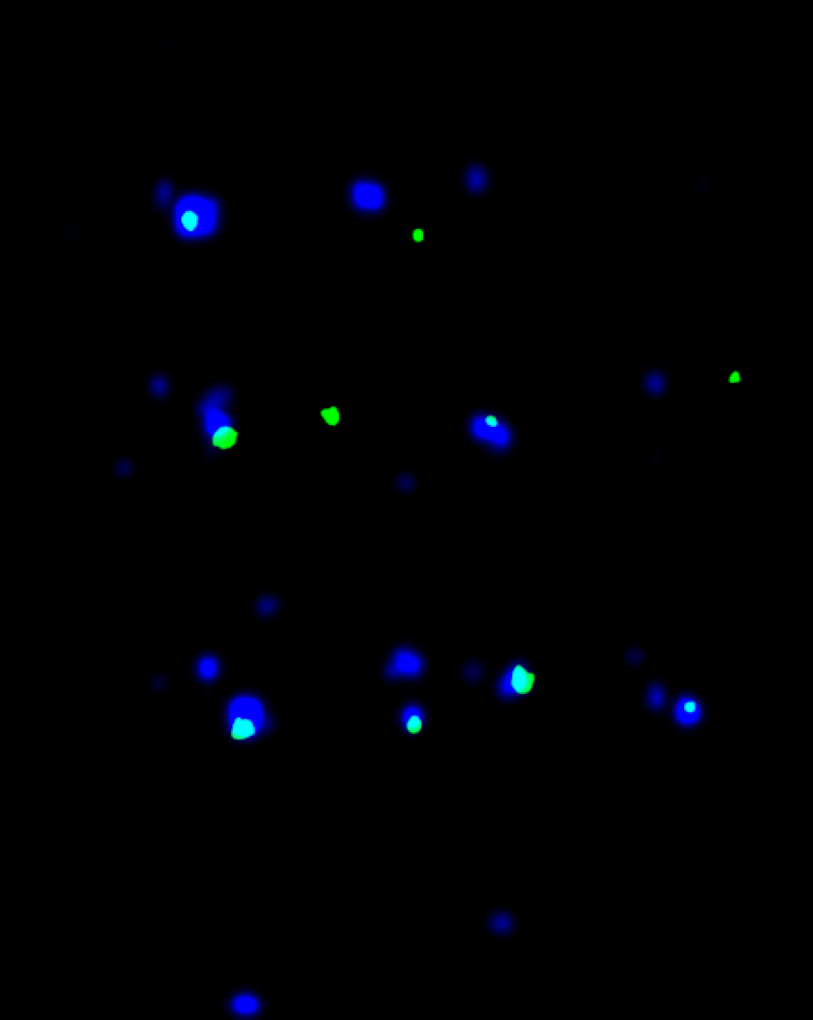}
         \caption{}
         \label{fig:resultsDiff8}
     \end{subfigure}
     \caption{Prediction example of log 8 compared to ground truth (a) raw height map (b) aligned RGB image and height map (c) predictions compared to ground truth (blue -- predictions, green -- ground truth, cyan -- correct predictions).}
     \label{fig:predExample8}
\end{figure}

The method fails on Log 6, which is very likely caused by the limited number of training samples, where Sample 6 represents the log type with properties that are not represented in other samples. Example predictions are shown in Fig.~\ref{fig:predExample1} and Fig.~\ref{fig:predExample8}. 

\begin{table}
\centering
\footnotesize
\caption{Knot detection results (mAP) for single-modality branches and two fusion approaches.}
\begin{tabularx}{1\columnwidth}
  {>{\centering\arraybackslash\hsize=0.4\hsize\columnwidth=\hsize}X
   | >{\centering\arraybackslash\hsize=0.35\hsize\columnwidth=\hsize}X
   >{\centering\arraybackslash\hsize=0.35\hsize\columnwidth=\hsize}X
   >{\centering\arraybackslash\hsize=0.45\hsize\columnwidth=\hsize}X
   >{\centering\arraybackslash\hsize=0.55\hsize\columnwidth=\hsize}X
  }

 &  &  & Fusion & Fusion \\[3pt]
Log & Height & Image & (max) & (proposed) \\[3pt]
\hline\\[-5pt]
1 & {0.82} & {0.40} & 0.38 & \textbf{0.90} \\ 
2 & {0.20} & {0.15} & 0.10 & \textbf{0.36} \\ 
3 & 0.50 & {0.18} & 0.16 & \textbf{0.56} \\ 
4 & {0.20} & {0.11} & 0.09 & \textbf{0.67} \\ 
5 & {0.20} & {0.04} & 0.04 & \textbf{0.67} \\ 
6 & 0.08 & 0.00 & 0.04 & {0.0} \\ 
7 & 0.39 & 0.26 & 0.23 & \textbf{0.64} \\ 
8 & \textbf{0.57} & {0.09} & 0.10 & \underline{0.54} \\ 
9 & 0.25 & 0.22 & 0.25 & \textbf{0.50} \\[2pt] 
\hline\\[-5pt]
All & \underline{0.36} & 0.16 & 0.15 & \textbf{0.54}\\
\end{tabularx}
\label{tab:detMet}
\end{table}

\renewcommand\tabularxcolumn[1]{m{#1}}
\begin{table*}[t]
\centering
\footnotesize
\caption{Comparison of sawing optimization results for different knot detection methods. The table is divided by the criteria, the first two rows show results for the quantity of arris knots, and the second part shows results for the arris knot area around the edges.
The first rows of each part indicate the improvements compared to random angle sawing (average results over all angles).}
\begin{tabularx}{0.9\textwidth}
  {>{\centering\arraybackslash\hsize=0.8\hsize\columnwidth=\hsize}X
   | >{\centering\arraybackslash\hsize=0.5\hsize\columnwidth=\hsize}X
   >{\centering\arraybackslash\hsize=0.5\hsize\columnwidth=\hsize}X
   >{\centering\arraybackslash\hsize=0.5\hsize\columnwidth=\hsize}X
   >{\centering\arraybackslash\hsize=0.5\hsize\columnwidth=\hsize}X
  }

Metric & Height map-based optimization & Image-based optimization & Maximum fusion optimization & Proposed fusion optimization \\[10pt]
\hline\\
Method arris knot count against avg. arris knot count $\downarrow$ & -1.8\% & -5.8\% & \underline{-11.3\%} & \textbf{-23.4\% }\\[13pt]
Method arris knot count against all knot types count $\downarrow$  & $27.9\%$ ($106 / 380$) & $25.6\%$ ($101 / 394$) &  \underline{25.3\%} ($97 / 383$) & \textbf{20.8\%} ($82 / 394$)\\[10pt]
\hline\\[-5pt]
Method arris knot area against avg. knot area $\downarrow$  & -4.6\% & \underline{-9.6\%} & -7.8\% & \textbf{-31.4\%}\\[11pt]
Total arris knot area [$dm^2$] $\downarrow$  & 239.9 & \underline{231.8} & 233.5 & \textbf{173.6}\\
\end{tabularx}
\label{tab:sawMet}
\end{table*}

\subsection{Sawing angle optimization}
\label{sec:sawOptim}
While the proposed knot detection method fails to detect a notable number of knots that are not clearly visible on the log surface, it is safe to assume that the detection results are accurate enough to be useful in sawing optimization. To test this, we applied the proposed sawing optimization algorithm using the knot detections as input and measured the number and total area of arris knots. These numbers were obtained by virtually sawing the CT reconstructions at the sawing angle produced by the sawing optimization method. Virtual sawing was performed by interpolating the slices corresponding to the board surfaces.

To demonstrate that the knot detection and sawing optimization methods improve board quality (by reducing the number of arris knots), we compared the pipeline with the average number and total area of arris knots over all possible sawing angles. The average values can be seen as expected values when a random sawing angle is used. The results are shown in Table~\ref{tab:sawMet}. As can be seen, all knot detection methods reduce the number and area of arris knots when the sawing angle is optimized, with the proposed method being the most effective. The proposed methods for knot detection and sawing angle optimization reduce the number of unwanted arris knots by 23\% and the total area by 31\% compared to a random sawing angle.

\section{Conclusion}
\label{sec:conclusion}
In this paper, we proposed a novel multimodal method for detecting knots on log surfaces. The method utilizes a set of RGB images and laser point clouds of the log surface as input, converting them into a stitched RGB map and a surface height map, respectively, which are then processed in separate branches. Finally, the feature maps from both branches are combined in a data fusion module to produce a segmentation mask for the knots. To demonstrate the utility of the detected knots, we further proposed a simple sawing optimization method to estimate the sawing angle that minimizes arris knots in the resulting boards. This method employs a fast correlation-based technique to find an optimal sawing angle without the need to exhaustively test different angles. The results show that the proposed knot detection method, combined with the sawing optimization, reduces the number of undesired arris knots by 23\% and their total area by 31\%. Being able to control the sawing optimization using only surface information suggests that log measurement systems do not need to rely on expensive CT scanners to be efficient for sawing optimization purposes.

\bibliographystyle{model5-names} 
\bibliography{bibliography}

\begin{thebibliography}{46}
\expandafter\ifx\csname natexlab\endcsname\relax\def\natexlab#1{#1}\fi
\providecommand{\url}[1]{\texttt{#1}}
\providecommand{\href}[2]{#2}
\providecommand{\path}[1]{#1}
\providecommand{\DOIprefix}{doi:}
\providecommand{\ArXivprefix}{arXiv:}
\providecommand{\URLprefix}{URL: }
\providecommand{\Pubmedprefix}{pmid:}
\providecommand{\doi}[1]{\href{https://doi.org/#1}{\path{#1}}}
\providecommand{\Pubmed}[1]{\href{pmid:#1}{\path{#1}}}
\providecommand{\bibinfo}[2]{#2}
\ifx\xfnm\relax \def\xfnm[#1]{\unskip,\space#1}\fi
\bibitem[{Batrakhanov et~al.(2021)Batrakhanov, Zolotarev, Eerola, Lensu \& K{\"a}lvi{\"a}inen}]{batrakhanov2021virtual}
\bibinfo{author}{Batrakhanov, D.}, \bibinfo{author}{Zolotarev, F.}, \bibinfo{author}{Eerola, T.}, \bibinfo{author}{Lensu, L.}, \& \bibinfo{author}{K{\"a}lvi{\"a}inen, H.} (\bibinfo{year}{2021}).
\newblock \bibinfo{title}{Virtual sawing using generative adversarial networks}.
\newblock In {\it \bibinfo{booktitle}{International Conference on Image and Vision Computing New Zealand}\/}.
\bibitem[{Besl \& McKay(1992)}]{ICP}
\bibinfo{author}{Besl, P.}, \& \bibinfo{author}{McKay, N.~D.} (\bibinfo{year}{1992}).
\newblock \bibinfo{title}{A method for registration of 3-d shapes}.
\newblock {\it \bibinfo{journal}{IEEE Transactions on Pattern Analysis and Machine Intelligence}\/},  {\it \bibinfo{volume}{14}\/}, \bibinfo{pages}{239--256}. \DOIprefix\doi{10.1109/34.121791}.
\bibitem[{Chang \& Gazo(2009)}]{chang}
\bibinfo{author}{Chang, S.~J.}, \& \bibinfo{author}{Gazo, R.} (\bibinfo{year}{2009}).
\newblock \bibinfo{title}{Measuring the effect of internal log defect scanning on the value of lumber produced}.
\newblock {\it \bibinfo{journal}{Forest Products Journal}\/},  {\it \bibinfo{volume}{59}\/}. \DOIprefix\doi{https://doi.org/10.13073/0015-7473-59.11.56}.
\bibitem[{Chen et~al.(2023)Chen, Wan, Song \& Liu}]{chen20233d}
\bibinfo{author}{Chen, G.}, \bibinfo{author}{Wan, L.}, \bibinfo{author}{Song, L.}, \& \bibinfo{author}{Liu, Z.} (\bibinfo{year}{2023}).
\newblock \bibinfo{title}{3d {{Perception Algorithm}} of {{Random Environment Based}} on {{Point Cloud Enhanced Rgb Fusion}}}.
\newblock \DOIprefix\doi{10.2139/ssrn.4600273}.
\bibitem[{Deng et~al.(2009)Deng, Dong, Socher, Li, Li \& Fei-Fei}]{imagenet}
\bibinfo{author}{Deng, J.}, \bibinfo{author}{Dong, W.}, \bibinfo{author}{Socher, R.}, \bibinfo{author}{Li, L.-J.}, \bibinfo{author}{Li, K.}, \& \bibinfo{author}{Fei-Fei, L.} (\bibinfo{year}{2009}).
\newblock \bibinfo{title}{Imagenet: A large-scale hierarchical image database}.
\newblock In {\it \bibinfo{booktitle}{2009 IEEE Conference on Computer Vision and Pattern Recognition}\/} (pp. \bibinfo{pages}{248--255}).
\newblock \DOIprefix\doi{10.1109/CVPR.2009.5206848}.
\bibitem[{Farin(1986)}]{cloughTocher}
\bibinfo{author}{Farin, G.} (\bibinfo{year}{1986}).
\newblock \bibinfo{title}{Triangular bernstein-bézier patches}.
\newblock {\it \bibinfo{journal}{Computer Aided Geometric Design}\/},  {\it \bibinfo{volume}{3}\/}, \bibinfo{pages}{83--127}. \DOIprefix\doi{https://doi.org/10.1016/0167-8396(86)90016-6}.
\bibitem[{Fredriksson(2014)}]{fredriksson}
\bibinfo{author}{Fredriksson, M.} (\bibinfo{year}{2014}).
\newblock \bibinfo{title}{Log sawing position optimization using computed tomography scanning}.
\newblock {\it \bibinfo{journal}{Wood Material Science \& Engineering}\/},  {\it \bibinfo{volume}{9}\/}, \bibinfo{pages}{110--119}. \DOIprefix\doi{https://doi.org/10.1080/17480272.2014.904430}.
\bibitem[{Giovannini et~al.(2019)Giovannini, Boschetto, Vicario, Cossi, Busatto, Ghidoni \& Ursella}]{giovanniniImproving}
\bibinfo{author}{Giovannini, S.}, \bibinfo{author}{Boschetto, D.}, \bibinfo{author}{Vicario, E.}, \bibinfo{author}{Cossi, M.}, \bibinfo{author}{Busatto, A.}, \bibinfo{author}{Ghidoni, S.}, \& \bibinfo{author}{Ursella, E.} (\bibinfo{year}{2019}).
\newblock \bibinfo{title}{Improving knot segmentation using {{Deep Learning}} techniques}.
\newblock {\it \bibinfo{journal}{Proceedings, 21st international nondestructive testing and evaluation of wood symposium}\/}, .
\bibitem[{Howard et~al.(2019)Howard, Sandler, Chen, Wang, Chen, Tan, Chu, Vasudevan, Zhu, Pang, Adam \& Le}]{MobileNetv3}
\bibinfo{author}{Howard, A.}, \bibinfo{author}{Sandler, M.}, \bibinfo{author}{Chen, B.}, \bibinfo{author}{Wang, W.}, \bibinfo{author}{Chen, L.-C.}, \bibinfo{author}{Tan, M.}, \bibinfo{author}{Chu, G.}, \bibinfo{author}{Vasudevan, V.}, \bibinfo{author}{Zhu, Y.}, \bibinfo{author}{Pang, R.}, \bibinfo{author}{Adam, H.}, \& \bibinfo{author}{Le, Q.} (\bibinfo{year}{2019}).
\newblock \bibinfo{title}{Searching for mobilenetv3}.
\newblock In {\it \bibinfo{booktitle}{2019 IEEE/CVF International Conference on Computer Vision (ICCV)}\/} (pp. \bibinfo{pages}{1314--1324}).
\newblock \DOIprefix\doi{10.1109/ICCV.2019.00140}.
\bibitem[{Khazem et~al.(2023)Khazem, Richard, Fix \& Pradalier}]{khazem2023Deep}
\bibinfo{author}{Khazem, S.}, \bibinfo{author}{Richard, A.}, \bibinfo{author}{Fix, J.}, \& \bibinfo{author}{Pradalier, C.} (\bibinfo{year}{2023}).
\newblock \bibinfo{title}{Deep learning for the detection of semantic features in tree {{X-ray CT}} scans}.
\newblock {\it \bibinfo{journal}{Artificial Intelligence in Agriculture}\/},  {\it \bibinfo{volume}{7}\/}, \bibinfo{pages}{13--26}. \DOIprefix\doi{10.1016/j.aiia.2022.12.001}.
\bibitem[{Kr{\"a}henb{\"u}hl et~al.(2013)Kr{\"a}henb{\"u}hl, Kerautret \& {Debled-Rennesson}}]{krahenbuhl2013Knot}
\bibinfo{author}{Kr{\"a}henb{\"u}hl, A.}, \bibinfo{author}{Kerautret, B.}, \& \bibinfo{author}{{Debled-Rennesson}, I.} (\bibinfo{year}{2013}).
\newblock \bibinfo{title}{Knot {{Segmentation}} in {{Noisy 3D Images}} of {{Wood}}}.
\newblock In \bibinfo{editor}{D.~Hutchison}, \bibinfo{editor}{T.~Kanade}, \bibinfo{editor}{J.~Kittler}, \bibinfo{editor}{J.~M. Kleinberg}, \bibinfo{editor}{F.~Mattern}, \bibinfo{editor}{J.~C. Mitchell}, \bibinfo{editor}{M.~Naor}, \bibinfo{editor}{O.~Nierstrasz}, \bibinfo{editor}{C.~Pandu~Rangan}, \bibinfo{editor}{B.~Steffen}, \bibinfo{editor}{M.~Sudan}, \bibinfo{editor}{D.~Terzopoulos}, \bibinfo{editor}{D.~Tygar}, \bibinfo{editor}{M.~Y. Vardi}, \bibinfo{editor}{G.~Weikum}, \bibinfo{editor}{R.~{Gonzalez-Diaz}}, \bibinfo{editor}{M.-J. Jimenez}, \& \bibinfo{editor}{B.~Medrano} (Eds.), {\it \bibinfo{booktitle}{Discrete {{Geometry}} for {{Computer Imagery}}}\/} (pp. \bibinfo{pages}{383--394}).
\newblock \bibinfo{address}{Berlin, Heidelberg}: \bibinfo{publisher}{Springer Berlin Heidelberg} volume \bibinfo{volume}{7749}.
\newblock \DOIprefix\doi{10.1007/978-3-642-37067-0\_33}.
\bibitem[{Kr{\"a}henb{\"u}hl et~al.(2014)Kr{\"a}henb{\"u}hl, Kerautret, {Debled-Rennesson}, Mothe \& Longuetaud}]{krahenbuhl2014Knot}
\bibinfo{author}{Kr{\"a}henb{\"u}hl, A.}, \bibinfo{author}{Kerautret, B.}, \bibinfo{author}{{Debled-Rennesson}, I.}, \bibinfo{author}{Mothe, F.}, \& \bibinfo{author}{Longuetaud, F.} (\bibinfo{year}{2014}).
\newblock \bibinfo{title}{Knot segmentation in {{3D CT}} images of wet wood}.
\newblock {\it \bibinfo{journal}{Pattern Recognition}\/},  {\it \bibinfo{volume}{47}\/}, \bibinfo{pages}{3852--3869}. \DOIprefix\doi{10.1016/j.patcog.2014.05.015}.
\bibitem[{Kretschmer et~al.(2013)Kretschmer, Kirchner, Morhart \& Spiecker}]{kretschmer2013New}
\bibinfo{author}{Kretschmer, U.}, \bibinfo{author}{Kirchner, N.}, \bibinfo{author}{Morhart, C.}, \& \bibinfo{author}{Spiecker, H.} (\bibinfo{year}{2013}).
\newblock \bibinfo{title}{A new approach to assessing tree stem quality characteristics using terrestrial laser scans}.
\newblock {\it \bibinfo{journal}{Silva Fennica}\/},  {\it \bibinfo{volume}{47}\/}.
\bibitem[{Longuetaud et~al.(2012)Longuetaud, Mothe, Kerautret, Kr{\"a}henb{\"u}hl, Hory, Leban \& {Debled-Rennesson}}]{longuetaud2012Automatic}
\bibinfo{author}{Longuetaud, F.}, \bibinfo{author}{Mothe, F.}, \bibinfo{author}{Kerautret, B.}, \bibinfo{author}{Kr{\"a}henb{\"u}hl, A.}, \bibinfo{author}{Hory, L.}, \bibinfo{author}{Leban, J.~M.}, \& \bibinfo{author}{{Debled-Rennesson}, I.} (\bibinfo{year}{2012}).
\newblock \bibinfo{title}{Automatic knot detection and measurements from {{X-ray CT}} images of wood: {{A}} review and validation of an improved algorithm on softwood samples}.
\newblock {\it \bibinfo{journal}{Computers and Electronics in Agriculture}\/},  {\it \bibinfo{volume}{85}\/}, \bibinfo{pages}{77--89}. \DOIprefix\doi{10.1016/j.compag.2012.03.013}.
\bibitem[{Lowe(2004)}]{lowe2004DistinctiveIF}
\bibinfo{author}{Lowe, D.~G.} (\bibinfo{year}{2004}).
\newblock \bibinfo{title}{Distinctive image features from scale-invariant keypoints}.
\newblock {\it \bibinfo{journal}{International Journal of Computer Vision}\/},  {\it \bibinfo{volume}{60}\/}, \bibinfo{pages}{91--110}. \DOIprefix\doi{https://doi.org/10.1023/B:VISI.0000029664.99615.94}.
\bibitem[{Lundahl \& Grönlund(2010)}]{lundahl}
\bibinfo{author}{Lundahl, C.~G.}, \& \bibinfo{author}{Grönlund, A.} (\bibinfo{year}{2010}).
\newblock \bibinfo{title}{Increased yield in sawmills by applying alternate rotation and lateral positioning}.
\newblock {\it \bibinfo{journal}{Forest Products Journal}\/},  {\it \bibinfo{volume}{60}\/}. \DOIprefix\doi{https://doi.org/10.13073/0015-7473-60.4.331}.
\bibitem[{Ma et~al.(2019)Ma, Wang, Li, Zhang, Ouyang \& Fan}]{ma2019Accurate}
\bibinfo{author}{Ma, X.}, \bibinfo{author}{Wang, Z.}, \bibinfo{author}{Li, H.}, \bibinfo{author}{Zhang, P.}, \bibinfo{author}{Ouyang, W.}, \& \bibinfo{author}{Fan, X.} (\bibinfo{year}{2019}).
\newblock \bibinfo{title}{Accurate {{Monocular 3D Object Detection}} via {{Color-Embedded 3D Reconstruction}} for {{Autonomous Driving}}}.
\newblock In {\it \bibinfo{booktitle}{2019 {{IEEE}}/{{CVF International Conference}} on {{Computer Vision}} ({{ICCV}})}\/} (pp. \bibinfo{pages}{6850--6859}).
\newblock \bibinfo{address}{Seoul, Korea (South)}: \bibinfo{publisher}{IEEE}.
\newblock \DOIprefix\doi{10.1109/ICCV.2019.00695}.
\bibitem[{Madawy et~al.(2019)Madawy, Rashed, Sallab, Nasr, Kamel \& Yogamani}]{madawy2019RGB}
\bibinfo{author}{Madawy, K.~E.}, \bibinfo{author}{Rashed, H.}, \bibinfo{author}{Sallab, A.~E.}, \bibinfo{author}{Nasr, O.}, \bibinfo{author}{Kamel, H.}, \& \bibinfo{author}{Yogamani, S.} (\bibinfo{year}{2019}).
\newblock \bibinfo{title}{{{RGB}} and {{LiDAR}} fusion based {{3D Semantic Segmentation}} for {{Autonomous Driving}}}.
\bibitem[{Norlander et~al.(2015)Norlander, Grahn \& Maki}]{norlander2015Wooden}
\bibinfo{author}{Norlander, R.}, \bibinfo{author}{Grahn, J.}, \& \bibinfo{author}{Maki, A.} (\bibinfo{year}{2015}).
\newblock \bibinfo{title}{Wooden {{Knot Detection Using ConvNet Transfer Learning}}}.
\newblock In \bibinfo{editor}{R.~R. Paulsen}, \& \bibinfo{editor}{K.~S. Pedersen} (Eds.), {\it \bibinfo{booktitle}{Image {{Analysis}}}\/} (pp. \bibinfo{pages}{263--274}).
\newblock \bibinfo{address}{Cham}: \bibinfo{publisher}{Springer International Publishing}.
\newblock \DOIprefix\doi{10.1007/978-3-319-19665-7\_22}.
\bibitem[{Qi et~al.(2017{\natexlab{a}})Qi, Su, Mo \& Guibas}]{qi2017PointNet}
\bibinfo{author}{Qi, C.~R.}, \bibinfo{author}{Su, H.}, \bibinfo{author}{Mo, K.}, \& \bibinfo{author}{Guibas, L.~J.} (\bibinfo{year}{2017}{\natexlab{a}}).
\newblock \bibinfo{title}{{{PointNet}}: {{Deep Learning}} on {{Point Sets}} for {{3D Classification}} and {{Segmentation}}}.
\newblock \DOIprefix\doi{10.48550/arXiv.1612.00593}.
\bibitem[{Qi et~al.(2017{\natexlab{b}})Qi, Yi, Su \& Guibas}]{qi2017PointNeta}
\bibinfo{author}{Qi, C.~R.}, \bibinfo{author}{Yi, L.}, \bibinfo{author}{Su, H.}, \& \bibinfo{author}{Guibas, L.~J.} (\bibinfo{year}{2017}{\natexlab{b}}).
\newblock \bibinfo{title}{{{PointNet}}++: {{Deep Hierarchical Feature Learning}} on {{Point Sets}} in a {{Metric Space}}}.
\newblock \bibinfo{howpublished}{https://arxiv.org/abs/1706.02413v1}.
\bibitem[{Rais et~al.(2017)Rais, Ursella, Vicario \& Giudiceandrea}]{rais}
\bibinfo{author}{Rais, A.}, \bibinfo{author}{Ursella, E.}, \bibinfo{author}{Vicario, E.}, \& \bibinfo{author}{Giudiceandrea, F.} (\bibinfo{year}{2017}).
\newblock \bibinfo{title}{The use of the first industrial x-ray ct scanner increases the lumber recovery value: case study on visually strength-graded douglas-fir timber}.
\newblock {\it \bibinfo{journal}{Annals of Forest Science}\/},  {\it \bibinfo{volume}{74}\/}. \DOIprefix\doi{https://doi.org/10.1007/s13595-017-0630-5}.
\bibitem[{Ren et~al.(2015)Ren, He, Girshick \& Sun}]{NIPS2015_14bfa6bb}
\bibinfo{author}{Ren, S.}, \bibinfo{author}{He, K.}, \bibinfo{author}{Girshick, R.}, \& \bibinfo{author}{Sun, J.} (\bibinfo{year}{2015}).
\newblock \bibinfo{title}{Faster r-cnn: Towards real-time object detection with region proposal networks}.
\newblock In \bibinfo{editor}{C.~Cortes}, \bibinfo{editor}{N.~Lawrence}, \bibinfo{editor}{D.~Lee}, \bibinfo{editor}{M.~Sugiyama}, \& \bibinfo{editor}{R.~Garnett} (Eds.), {\it \bibinfo{booktitle}{Advances in Neural Information Processing Systems}\/}.
\newblock \bibinfo{publisher}{Curran Associates, Inc.} volume~\bibinfo{volume}{28}.
\bibitem[{Riegler et~al.(2017)Riegler, Ulusoy \& Geiger}]{riegler2017OctNet}
\bibinfo{author}{Riegler, G.}, \bibinfo{author}{Ulusoy, A.~O.}, \& \bibinfo{author}{Geiger, A.} (\bibinfo{year}{2017}).
\newblock \bibinfo{title}{{{OctNet}}: {{Learning Deep 3D Representations}} at {{High Resolutions}}}.
\newblock \DOIprefix\doi{10.48550/arXiv.1611.05009}.
\bibitem[{Ruz et~al.(2005)Ruz, Est{\'e}vez \& Perez}]{ruzNeurofuzzy}
\bibinfo{author}{Ruz, G.~A.}, \bibinfo{author}{Est{\'e}vez, P.~A.}, \& \bibinfo{author}{Perez, C.~A.} (\bibinfo{year}{2005}).
\newblock \bibinfo{title}{A neurofuzzy color image segmentation method for wood surface defect detection}.
\newblock {\it \bibinfo{journal}{Forest Products Journal}\/}, .
\bibitem[{{Salazar-Gomez} et~al.(2022){Salazar-Gomez}, {Sierra-Gonzalez}, {Diaz-Zapata}, Paigwar, Liu, Erkent \& Laugier}]{salazar-gomez2022TransFuseGrid}
\bibinfo{author}{{Salazar-Gomez}, G.}, \bibinfo{author}{{Sierra-Gonzalez}, D.}, \bibinfo{author}{{Diaz-Zapata}, M.}, \bibinfo{author}{Paigwar, A.}, \bibinfo{author}{Liu, W.}, \bibinfo{author}{Erkent, O.}, \& \bibinfo{author}{Laugier, C.} (\bibinfo{year}{2022}).
\newblock \bibinfo{title}{{{TransFuseGrid}}: {{Transformer-based Lidar-RGB}} fusion for semantic grid prediction}.
\newblock In {\it \bibinfo{booktitle}{2022 17th {{International Conference}} on {{Control}}, {{Automation}}, {{Robotics}} and {{Vision}} ({{ICARCV}})}\/} (pp. \bibinfo{pages}{268--273}).
\newblock \bibinfo{address}{Singapore, Singapore}: \bibinfo{publisher}{IEEE}.
\newblock \DOIprefix\doi{10.1109/ICARCV57592.2022.10004276}.
\bibitem[{Seferbekov et~al.(2018)Seferbekov, Iglovikov, Buslaev \& Shvets}]{FPNSegmentation}
\bibinfo{author}{Seferbekov, S.}, \bibinfo{author}{Iglovikov, V.}, \bibinfo{author}{Buslaev, A.}, \& \bibinfo{author}{Shvets, A.} (\bibinfo{year}{2018}).
\newblock \bibinfo{title}{Feature pyramid network for multi-class land segmentation}.
\newblock In {\it \bibinfo{booktitle}{2018 IEEE/CVF Conference on Computer Vision and Pattern Recognition Workshops (CVPRW)}\/} (pp. \bibinfo{pages}{272--2723}).
\newblock \DOIprefix\doi{10.1109/CVPRW.2018.00051}.
\bibitem[{Stängle et~al.(2015)Stängle, Brüchert, Heikkila, Usenius, Usenius \& Sauter}]{stangle}
\bibinfo{author}{Stängle, S.~M.}, \bibinfo{author}{Brüchert, F.}, \bibinfo{author}{Heikkila, A.}, \bibinfo{author}{Usenius, T.}, \bibinfo{author}{Usenius, A.}, \& \bibinfo{author}{Sauter, U.~H.} (\bibinfo{year}{2015}).
\newblock \bibinfo{title}{Potentially increased sawmill yield from hardwoods using x-ray computed tomography for knot detection}.
\newblock {\it \bibinfo{journal}{Annals of Forest Science}\/},  {\it \bibinfo{volume}{72}\/}. \DOIprefix\doi{https://doi.org/10.1007/s13595-014-0385-1}.
\bibitem[{Thomas \& Mili(2007)}]{thomas2007Robust}
\bibinfo{author}{Thomas, L.}, \& \bibinfo{author}{Mili, L.} (\bibinfo{year}{2007}).
\newblock \bibinfo{title}{A {{Robust GM-Estimator}} for the {{Automated Detection}} of {{External Defects}} on {{Barked Hardwood Logs}} and {{Stems}}}.
\newblock {\it \bibinfo{journal}{IEEE Transactions on Signal Processing}\/},  {\it \bibinfo{volume}{55}\/}, \bibinfo{pages}{3568--3576}. \DOIprefix\doi{10.1109/TSP.2007.894262}.
\bibitem[{Thomas et~al.(2007)Thomas, Shaffer, Mili \& Thomas}]{thomas2007Automated}
\bibinfo{author}{Thomas, L.}, \bibinfo{author}{Shaffer, C.}, \bibinfo{author}{Mili, L.}, \& \bibinfo{author}{Thomas, E.} (\bibinfo{year}{2007}).
\newblock \bibinfo{title}{Automated detection of severe surface defects on barked hardwood logs}.
\newblock {\it \bibinfo{journal}{Forest {Products} {Journal}}\/},  {\it \bibinfo{volume}{57}\/}.
\bibitem[{Thomas \& Thomas(2011)}]{thomas2011Graphical}
\bibinfo{author}{Thomas, L.}, \& \bibinfo{author}{Thomas, R.~E.} (\bibinfo{year}{2011}).
\newblock \bibinfo{title}{A graphical automated detection system to locate hardwood log surface defects using high-resolution three-dimensional laser scan data}.
\newblock {\it \bibinfo{journal}{In: Fei, Songlin; Lhotka, John M.; Stringer, Jeffrey W.; Gottschalk, Kurt W.; Miller, Gary W., eds. Proceedings, 17th central hardwood forest conference; 2010 April 5-7; Lexington, KY; Gen. Tech. Rep. NRS-P-78. Newtown Square, PA: U.S. Department of Agriculture, Forest Service, Northern Research Station: 92-101.}\/},  {\it \bibinfo{volume}{78}\/}, \bibinfo{pages}{92--101}.
\bibitem[{Todoroki et~al.(2010)Todoroki, Lowell \& Dykstra}]{todoroki2010Automated}
\bibinfo{author}{Todoroki, C.}, \bibinfo{author}{Lowell, E.}, \& \bibinfo{author}{Dykstra, D.} (\bibinfo{year}{2010}).
\newblock \bibinfo{title}{Automated knot detection with visual post-processing of {{Douglas-fir}} veneer images}.
\newblock {\it \bibinfo{journal}{Computers and Electronics in Agriculture}\/},  {\it \bibinfo{volume}{70}\/}, \bibinfo{pages}{163--171}. \DOIprefix\doi{10.1016/j.compag.2009.10.002}.
\bibitem[{Urbonas et~al.(2019)Urbonas, Raudonis, Maskeli{\=u}nas \& Dama{\v s}evi{\v c}ius}]{urbonas2019Automated}
\bibinfo{author}{Urbonas, A.}, \bibinfo{author}{Raudonis, V.}, \bibinfo{author}{Maskeli{\=u}nas, R.}, \& \bibinfo{author}{Dama{\v s}evi{\v c}ius, R.} (\bibinfo{year}{2019}).
\newblock \bibinfo{title}{Automated {{Identification}} of {{Wood Veneer Surface Defects Using Faster Region-Based Convolutional Neural Network}} with {{Data Augmentation}} and {{Transfer Learning}}}.
\newblock {\it \bibinfo{journal}{Applied Sciences}\/},  {\it \bibinfo{volume}{9}\/}, \bibinfo{pages}{4898}. \DOIprefix\doi{10.3390/app9224898}.
\bibitem[{{Van-Tho Nguyen} et~al.(2016){Van-Tho Nguyen}, Kerautret, {Debled-Rennesson}, Colin, Piboule \& Constant}]{van-thonguyen2016Segmentation}
\bibinfo{author}{{Van-Tho Nguyen}}, \bibinfo{author}{Kerautret, B.}, \bibinfo{author}{{Debled-Rennesson}, I.}, \bibinfo{author}{Colin, F.}, \bibinfo{author}{Piboule, A.}, \& \bibinfo{author}{Constant, T.} (\bibinfo{year}{2016}).
\newblock \bibinfo{title}{Segmentation of defects on log surface from terrestrial lidar data}.
\newblock In {\it \bibinfo{booktitle}{2016 23rd {{International Conference}} on {{Pattern Recognition}} ({{ICPR}})}\/} (pp. \bibinfo{pages}{3168--3173}).
\newblock \bibinfo{address}{Cancun}: \bibinfo{publisher}{IEEE}.
\newblock \DOIprefix\doi{10.1109/ICPR.2016.7900122}.
\bibitem[{Wang et~al.(2019)Wang, Sun, Liu, Sarma, Bronstein \& Solomon}]{wang2019Dynamic}
\bibinfo{author}{Wang, Y.}, \bibinfo{author}{Sun, Y.}, \bibinfo{author}{Liu, Z.}, \bibinfo{author}{Sarma, S.~E.}, \bibinfo{author}{Bronstein, M.~M.}, \& \bibinfo{author}{Solomon, J.~M.} (\bibinfo{year}{2019}).
\newblock \bibinfo{title}{Dynamic {{Graph CNN}} for {{Learning}} on {{Point Clouds}}}.
\newblock \DOIprefix\doi{10.48550/arXiv.1801.07829}.
\bibitem[{Wu et~al.(2023)Wu, Jiang, Wang, Liu, Liu, Qiao, Ouyang, He \& Zhao}]{wu2023Point}
\bibinfo{author}{Wu, X.}, \bibinfo{author}{Jiang, L.}, \bibinfo{author}{Wang, P.-S.}, \bibinfo{author}{Liu, Z.}, \bibinfo{author}{Liu, X.}, \bibinfo{author}{Qiao, Y.}, \bibinfo{author}{Ouyang, W.}, \bibinfo{author}{He, T.}, \& \bibinfo{author}{Zhao, H.} (\bibinfo{year}{2023}).
\newblock \bibinfo{title}{Point {{Transformer V3}}: {{Simpler}}, {{Faster}}, {{Stronger}}}.
\newblock \bibinfo{howpublished}{https://arxiv.org/abs/2312.10035v2}.
\bibitem[{Wu et~al.(2022)Wu, Lao, Jiang, Liu \& Zhao}]{wu2022Point}
\bibinfo{author}{Wu, X.}, \bibinfo{author}{Lao, Y.}, \bibinfo{author}{Jiang, L.}, \bibinfo{author}{Liu, X.}, \& \bibinfo{author}{Zhao, H.} (\bibinfo{year}{2022}).
\newblock \bibinfo{title}{Point {{Transformer V2}}: {{Grouped Vector Attention}} and {{Partition-based Pooling}}}.
\newblock \DOIprefix\doi{10.48550/arXiv.2210.05666}.
\bibitem[{Xu et~al.(2021)Xu, Ding, Zhao \& Qi}]{xu2021PAConv}
\bibinfo{author}{Xu, M.}, \bibinfo{author}{Ding, R.}, \bibinfo{author}{Zhao, H.}, \& \bibinfo{author}{Qi, X.} (\bibinfo{year}{2021}).
\newblock \bibinfo{title}{{{PAConv}}: {{Position Adaptive Convolution}} with {{Dynamic Kernel Assembling}} on {{Point Clouds}}}.
\newblock \DOIprefix\doi{10.48550/arXiv.2103.14635}.
\bibitem[{Yuan et~al.(2023)Yuan, Fu, Li, Meng \& Wang}]{yuan2023PointMBF}
\bibinfo{author}{Yuan, M.}, \bibinfo{author}{Fu, K.}, \bibinfo{author}{Li, Z.}, \bibinfo{author}{Meng, Y.}, \& \bibinfo{author}{Wang, M.} (\bibinfo{year}{2023}).
\newblock \bibinfo{title}{{{PointMBF}}: {{A Multi-scale Bidirectional Fusion Network}} for {{Unsupervised RGB-D Point Cloud Registration}}}.
\newblock In {\it \bibinfo{booktitle}{2023 {{IEEE}}/{{CVF International Conference}} on {{Computer Vision}} ({{ICCV}})}\/} (pp. \bibinfo{pages}{17648--17659}).
\newblock \bibinfo{address}{Paris, France}: \bibinfo{publisher}{IEEE}.
\newblock \DOIprefix\doi{10.1109/ICCV51070.2023.01622}.
\bibitem[{Zhang et~al.(2022)Zhang, Zhou, Liu, Sun, Chen \& Wang}]{Zhang_2023}
\bibinfo{author}{Zhang, W.}, \bibinfo{author}{Zhou, F.}, \bibinfo{author}{Liu, Y.}, \bibinfo{author}{Sun, P.}, \bibinfo{author}{Chen, Y.}, \& \bibinfo{author}{Wang, L.} (\bibinfo{year}{2022}).
\newblock \bibinfo{title}{Object defect detection based on data fusion of a 3d point cloud and 2d image}.
\newblock {\it \bibinfo{journal}{Measurement Science and Technology}\/},  {\it \bibinfo{volume}{34}\/}, \bibinfo{pages}{025002}. \DOIprefix\doi{10.1088/1361-6501/ac93a3}.
\bibitem[{Zhang et~al.(2023)Zhang, Jiang, Zhang \& Chen}]{zhang2023Threedimensional}
\bibinfo{author}{Zhang, Y.}, \bibinfo{author}{Jiang, D.}, \bibinfo{author}{Zhang, Z.}, \& \bibinfo{author}{Chen, J.} (\bibinfo{year}{2023}).
\newblock \bibinfo{title}{Three-dimensional inversion of knot defects recognition in timber cutting}.
\newblock {\it \bibinfo{journal}{Journal of Forestry Research}\/},  {\it \bibinfo{volume}{34}\/}, \bibinfo{pages}{1145--1152}. \DOIprefix\doi{10.1007/s11676-022-01532-y}.
\bibitem[{Zhang et~al.(2020)Zhang, Zhou, David, Yue, Xi, Gong \& Foroosh}]{zhang2020PolarNet}
\bibinfo{author}{Zhang, Y.}, \bibinfo{author}{Zhou, Z.}, \bibinfo{author}{David, P.}, \bibinfo{author}{Yue, X.}, \bibinfo{author}{Xi, Z.}, \bibinfo{author}{Gong, B.}, \& \bibinfo{author}{Foroosh, H.} (\bibinfo{year}{2020}).
\newblock \bibinfo{title}{{{PolarNet}}: {{An Improved Grid Representation}} for {{Online LiDAR Point Clouds Semantic Segmentation}}}.
\newblock \bibinfo{howpublished}{https://arxiv.org/abs/2003.14032v2}.
\bibitem[{Zhao et~al.(2020)Zhao, Jiang, Jia, Torr \& Koltun}]{zhao2020Point}
\bibinfo{author}{Zhao, H.}, \bibinfo{author}{Jiang, L.}, \bibinfo{author}{Jia, J.}, \bibinfo{author}{Torr, P.}, \& \bibinfo{author}{Koltun, V.} (\bibinfo{year}{2020}).
\newblock \bibinfo{title}{Point {{Transformer}}}.
\newblock \bibinfo{howpublished}{https://arxiv.org/abs/2012.09164v2}.
\bibitem[{Zhao et~al.(2022)Zhao, Ma, Meng, Liu, Wang, Junior, Gon{\c c}alves \& Li}]{zhao20223D}
\bibinfo{author}{Zhao, K.}, \bibinfo{author}{Ma, L.}, \bibinfo{author}{Meng, Y.}, \bibinfo{author}{Liu, L.}, \bibinfo{author}{Wang, J.}, \bibinfo{author}{Junior, J.}, \bibinfo{author}{Gon{\c c}alves, W.}, \& \bibinfo{author}{Li, J.} (\bibinfo{year}{2022}).
\newblock \bibinfo{title}{{{3D Vehicle Detection Using Multi-Level Fusion From Point Clouds}} and {{Images}}}.
\newblock {\it \bibinfo{journal}{IEEE Transactions on Intelligent Transportation Systems}\/},  {\it \bibinfo{volume}{PP}\/}. \DOIprefix\doi{10.1109/TITS.2021.3137392}.
\bibitem[{Zolotarev et~al.(2020{\natexlab{a}})Zolotarev, Eerola, Lensu, K{\"a}lvi{\"a}inen, Helin, Haario, Kauppi \& Heikkinen}]{zolotarev2020Modelling}
\bibinfo{author}{Zolotarev, F.}, \bibinfo{author}{Eerola, T.}, \bibinfo{author}{Lensu, L.}, \bibinfo{author}{K{\"a}lvi{\"a}inen, H.}, \bibinfo{author}{Helin, T.}, \bibinfo{author}{Haario, H.}, \bibinfo{author}{Kauppi, T.}, \& \bibinfo{author}{Heikkinen, J.} (\bibinfo{year}{2020}{\natexlab{a}}).
\newblock \bibinfo{title}{Modelling internal knot distribution using external log features}.
\newblock {\it \bibinfo{journal}{Computers and Electronics in Agriculture}\/},  {\it \bibinfo{volume}{179}\/}, \bibinfo{pages}{105795}. \DOIprefix\doi{10.1016/j.compag.2020.105795}.
\bibitem[{Zolotarev et~al.(2020{\natexlab{b}})Zolotarev, Eerola, Lensu, Kälviäinen, Helin, Haario, Kauppi \& Heikkinen}]{zolotarevKnotModelling}
\bibinfo{author}{Zolotarev, F.}, \bibinfo{author}{Eerola, T.}, \bibinfo{author}{Lensu, L.}, \bibinfo{author}{Kälviäinen, H.}, \bibinfo{author}{Helin, T.}, \bibinfo{author}{Haario, H.}, \bibinfo{author}{Kauppi, T.}, \& \bibinfo{author}{Heikkinen, J.} (\bibinfo{year}{2020}{\natexlab{b}}).
\newblock \bibinfo{title}{Modelling internal knot distribution using external log features}.
\newblock {\it \bibinfo{journal}{Computers and Electronics in Agriculture}\/},  {\it \bibinfo{volume}{179}\/}, \bibinfo{pages}{105795}. \DOIprefix\doi{https://doi.org/10.1016/j.compag.2020.105795}.

\end{thebibliography}

\end{document}